\documentclass[preprint,12pt]{elsarticle}

\usepackage{amsmath,amssymb,amsfonts}
\usepackage{graphicx}
\usepackage{booktabs}
\usepackage[ruled,linesnumbered]{algorithm2e}
\usepackage{xspace}
\usepackage{float}
\usepackage{url}
\usepackage{chngcntr}
\usepackage{lineno}
\usepackage[colorlinks=true,linkcolor=blue,citecolor=blue,urlcolor=blue]{hyperref}

\biboptions{sort&compress}

\newcommand{\ours}{\emph{ours}\xspace}
\newcommand{\noapf}{\emph{no\textunderscore apf}\xspace}
\newcommand{\noprog}{\emph{no\textunderscore progress}\xspace}
\newcommand{\noxy}{\emph{no\textunderscore xy\textunderscore apf}\xspace}
\newcommand{\vanilla}{\emph{vanilla}\xspace}

\title{Positive Safety Margins from Artificial Potential Field Reward Augmentation for Size-Robust Obstacle Avoidance under Partial Observability}
\author[1]{Ziyi Hao\corref{cor1}\fnref{fn1}}
\cortext[cor1]{Corresponding author.}
\ead{1155237839@link.cuhk.edu.hk}
\fntext[fn1]{Present address: Tianjin, China.}
\affiliation[1]{organization={The Chinese University of Hong Kong},
            addressline={Shatin, New Territories},
            city={Hong Kong SAR},
            country={China}}

\begin{document}

\begin{abstract}
Collision-free reaching is central to manipulation, yet real
cells often leave the size and exact position of an obstacle
unobserved, which excludes reinforcement learning controllers that take obstacle
geometry as input.
A reward limited to a collision penalty does not depend on the clearance outside the
obstacle, so under a step cost its learned policy skims the obstacle surface and
fails once the obstacle radius exceeds its training value. We investigate whether a
single analytic reward term can induce a safety margin without exposing the obstacle
to the policy. We augment the reward of a simulated Franka Emika Panda reaching task with a horizontal artificial potential field that penalizes proximity
within a safety margin around each of two pillars (fixed nominal layout, random drift). Under hidden observation, policies trained with this term maintain a
safe success rate above $88\%$ across a radius sweep from $0.03\,\mathrm{m}$ to
$0.055\,\mathrm{m}$ ($30$ seeds), while policies without it fall below $10\%$ at the
maximum tested radius; the term induces a minimum clearance of $47.9\,\mathrm{mm}$ against
$9.6\,\mathrm{mm}$ without it and eliminates training collisions at least eight
times earlier (median steps). The advantage is statistically significant at
all out-of-distribution radii and holds across three reinforcement learning
solvers and under full observation. Its safe success rate is comparable to that of a
classical potential-field planner given the pillar positions. The margin is a fixed buffer rather than an adaptation to the radius, and
it lengthens the path by about $19\%$.
\end{abstract}
\maketitle
\section{Introduction}\label{sec:introduction}

Robotic manipulators are deployed across assembly, logistics, and human-robot
collaboration, where collision-free reaching is a prerequisite for safe and
reliable operation. Real manipulation cells, however, seldom expose their
surroundings fully to the controller. Occlusion by the arm itself or by other
objects, together with noisy pose estimates, can remove obstacles from the
observation stream~\cite{garrett2020}, leaving the
controller to act while the exact position, motion, and size of the obstacle are
unknown. In this setting the task becomes partially
observable~\cite{pajarinen2017}, and the policy cannot acquire avoidance
behavior through perception alone. We study a fixed workcell: the nominal
obstacle layout is the same in every episode and can therefore be learned
implicitly from experience, whereas the obstacle size and the instantaneous
position, which drifts around the nominal one, are not observed. We call this
protocol \emph{hidden observation}.

Existing approaches fall short of this setting in three respects. Methods
that plan in a fully observed model assume the obstacle geometry is available,
either to a planner or to the policy. Recent reinforcement learning controllers
for manipulator obstacle avoidance follow this assumption: they include the
obstacle distance in the state observation~\cite{wang2025} or derive the
reward from an artificial potential field whose attraction and repulsion are
computed from the known obstacle positions~\cite{hu2024,xu2024,schneider2025}. These
methods do not ask whether the learned behavior survives a change in obstacle
size. Safe reinforcement learning methods that constrain a cost
signal~\cite{brunke2022,gu2024,garcia2015} are not excluded by hidden observation,
because a cost reaches the policy through training in the same way as a reward; they
bound an expected cost, however, rather than prescribe a clearance, and filters
that act on the obstacle state or model at run time need information that hidden
observation withholds (Section~\ref{sec:related-safe}). A
classical artificial potential field used directly as a
planner~\cite{khatib1986} requires no learning, but it needs the obstacle
positions and radius at run time, which are exactly what the hidden-observation
setting withholds (Section~\ref{sec:exp-classic}). To our knowledge, none of these lines of work
addresses the combination we study: a policy that does not observe the obstacle
size or its instantaneous position, evaluated against obstacle sizes beyond those
seen in training.

The failure that this combination induces is specific and structural. A reward
that penalizes only collision is flat with respect to clearance everywhere
outside the obstacle, so under a per-step cost the minimum-cost policy should
converge to zero clearance, resting on the obstacle surface or penetrating it
wherever penetration is cheaper than a detour. Such a policy is valid within
the training distribution, but it stores the obstacle boundary as a learned
absolute distance from the pillar center; the center is learnable because the
nominal layout is fixed, whereas the radius is not observed. When the obstacle radius exceeds its
training value, a trajectory that was barely collision-free begins to penetrate
the obstacle, and the policy has no signal with which to detect the change.

We address this failure by injecting an analytic prior into the reward. We
augment the dense reaching reward with a horizontal artificial potential field
that penalizes proximity to each obstacle within a safety margin around its
surface, while the obstacle itself remains unobserved. The term is a reward
augmentation rather than a potential-based shaping: it is a state-dependent
cost, not a potential difference, so the policy-invariance guarantee of
potential-based shaping~\cite{ng1999} does not apply and the term changes the
optimal policy by construction. It moves the penalty boundary from the obstacle
surface outward, to the safety margin, so the learned optimum becomes a
positive safety margin rather than contact with the surface. Because the policy
never observes the radius, the induced margin is a fixed buffer at the training
radius rather than an adaptation to the test-time radius. The buffer is
$47.9\,\mathrm{mm}$ wide, which exceeds the largest radius increase tested here
($25\,\mathrm{mm}$), so the prior makes the behavior robust to moderate changes of
a quantity that the policy cannot observe.

This paper makes three contributions. First, we demonstrate the near-zero-margin
failure under hidden observation: without the APF term, the policy
converges to a minimum clearance of $9.6\,\mathrm{mm}$ at the training radius,
and its safe success rate collapses as the radius grows. Second, we show that a
single horizontal APF term induces a positive safety margin of
$47.9\,\mathrm{mm}$ and maintains the safe success rate, the fraction of
episodes that reach the goal without collision, above $88\%$ across the entire
radius sweep from $0.03\,\mathrm{m}$ to $0.055\,\mathrm{m}$, while eliminating
collisions during training at least eight times earlier in median training steps,
with the advantage statistically significant at all out-of-distribution radii,
and that this margin is specific to the potential: raising the collision penalty
$25$--$50\times$ produces a margin in only $1$--$2$ of $10$ seeds.
Third, we show that the effect is not specific to one configuration: it persists
under full observation and across three RL solvers, and the resulting policy,
which does not observe the obstacles, reaches a safe success rate comparable to that
of a classical APF planner that reads the pillar positions.
The evaluation is limited to one simulated reaching task, and the margin is fixed
rather than adaptive: the safe success rate declines at the largest radii, and
Section~\ref{sec:discussion} states the remaining limitations.

\section{Related Work}\label{sec:related}

This work lies at the intersection of three lines of research: reinforcement
learning for manipulator motion planning, reward shaping and reward
augmentation, and safe reinforcement learning.

\subsection{Reinforcement learning for manipulator motion planning}

Obstacle avoidance for manipulators dates to the artificial potential field of
Khatib~\cite{khatib1986}, and deep reinforcement learning has since been applied
extensively to trajectory planning and avoidance. Most of these methods give the
controller access to obstacle information. Wang and
Kasaei~\cite{wang2025} combine a vision-based path planner in task space with
proximal policy optimization for obstacle avoidance in joint space, where a
distance-based penalty in the reward encourages clearance and the state
includes the distances between the obstacles and the manipulator links; that
work targets fast, obstacle-free trajectory generation and does not examine
obstacle sizes beyond those met in training. A second
line of work uses the artificial potential field directly as a reward. Hu et
al.~\cite{hu2024} treat the attractive force as a positive reward and the
repulsive force as a negative reward within a soft actor-critic controller; Xu
et al.~\cite{xu2024} guide a TD3 policy with an APF-derived potential field for
rock-drilling arms; and Schneider et al.~\cite{schneider2025} integrate a
modified potential field into deep deterministic policy gradient, supplying the
resulting force direction to the actor and adding a compensation term that
corrects suboptimal motion directions. These methods report gains in convergence and path efficiency, yet
all of them compute the distance or potential-field terms from known obstacle
positions. To our knowledge, none treats the obstacle as unobserved at
execution time, and none probes whether the learned behavior survives an
out-of-distribution obstacle size. The present work differs on both counts: the
obstacle is withheld from the observation, and the evaluation sweeps the
obstacle radius beyond its training value. Under hidden observation, the reward is the only channel through which
information about the obstacle size reaches the policy, which is the setting we study.

\subsection{Reward shaping and reward augmentation}

Reward shaping adds auxiliary rewards to guide learning. The classical theory of
Ng, Harada, and Russell~\cite{ng1999} shows that a shaping term of the form
$F(s,s')=\gamma\Phi(s')-\Phi(s)$, the difference of a potential function
evaluated at two successive states, leaves the optimal policy unchanged, and,
without further assumptions on the decision process, this potential-based form
is the only shaping that is guaranteed to preserve it. Devlin and Kudenko~\cite{devlin2012} extend the guarantee to dynamic
potentials. These methods aim to accelerate learning without changing what is
learned. The term we propose is not of this form: it is a state-dependent cost,
$\Phi(\mathbf{p})$ in Equation~\eqref{eq:apf}, rather than a potential
difference, and it changes the optimal policy by construction, moving the
penalty boundary from the obstacle surface outward. We therefore refer to it as a reward
augmentation rather than potential-based shaping, and the distinction is
central to the claim: the goal is not only to speed up convergence to a fixed optimum
but to replace a near-zero-margin optimum with a positive-margin one.

\subsection{Safe reinforcement learning}\label{sec:related-safe}

Safe RL constrains the learned policy to satisfy safety requirements, most often
through constrained optimization or explicit safety filters. Garc\'ia and
Fern\'andez~\cite{garcia2015} classify safe RL into methods that modify the
optimization criterion and methods that guide exploration with external
knowledge. More recent surveys cover safe learning in
robotics~\cite{brunke2022} and the methods, theory, and applications of safe
RL~\cite{gu2024}. Constrained policy optimization~\cite{achiam2017} bounds an
expected cumulative cost during policy search, and PID Lagrangian
methods~\cite{stooke2020} stabilize the dual update that enforces such a bound.
These methods need only a scalar cost signal, which reaches the policy through
training in the same way as the reward, so hidden observation does not exclude
them. What they enforce, however, is a limit on the expected cost, which is
typically a collision indicator; the clearance outside the obstacle enters the
constraint only if the cost is designed to include it, and a policy that satisfies
the limit at the training radius is not thereby guaranteed a margin at larger radii.
We do not compare against these methods, and combining a Lagrangian method with a
clearance-dependent cost is a direction for future work. Other safeguards act on
the observed state or on an obstacle model at run time. Barrier-function
layers~\cite{cheng2019} correct the action using a barrier defined on the obstacle
state and a dynamics model, shielding~\cite{alshiekh2018} blocks actions that
violate a specification over the observed state, and Recovery
RL~\cite{thananjeyan2021} switches to a learned recovery policy in regions where
violation is likely. Under hidden observation these safeguards lack the obstacle
information on which they are defined, unless it is restored to the observation. An alternative that also
leaves the observation unchanged is domain randomization~\cite{tobin2017}, which
varies the obstacle radius during training. We do not use it, because the
evaluation here fixes the training radius and tests radii beyond it, a premise
that randomization over the radius would remove.The reward augmentation we study instead encodes the safety
prior at training time, so that a policy that observes only its own state and
the goal maintains a safety margin at execution time.


\section{Method}\label{sec:method}

We address safe reaching for a manipulator when the obstacles are not
observable. The core contribution of this method is a reward augmentation: an
artificial-potential-field term is added to the dense reaching reward to encode
an analytic prior for obstacle avoidance. Because the policy does not observe the
obstacles, the reward is the only channel through which information about their
size reaches it. The potential term supplies this signal before contact, and it
enlarges the clearance that the learned policy maintains from the pillars, so that
the clearance remains positive when the obstacle size exceeds its training value.
The section is organized as
follows. Section~\ref{sec:method-mdp} formalizes the task as a continuous
Markov decision process, and Section~\ref{sec:method-obs} defines the
observation model. Section~\ref{sec:method-reward} defines the reward, including
the artificial-potential-field augmentation, and Section~\ref{sec:method-groups}
defines the reward configurations used to attribute the effect.

\begin{figure}[!t]
\centering
\includegraphics[width=\linewidth]{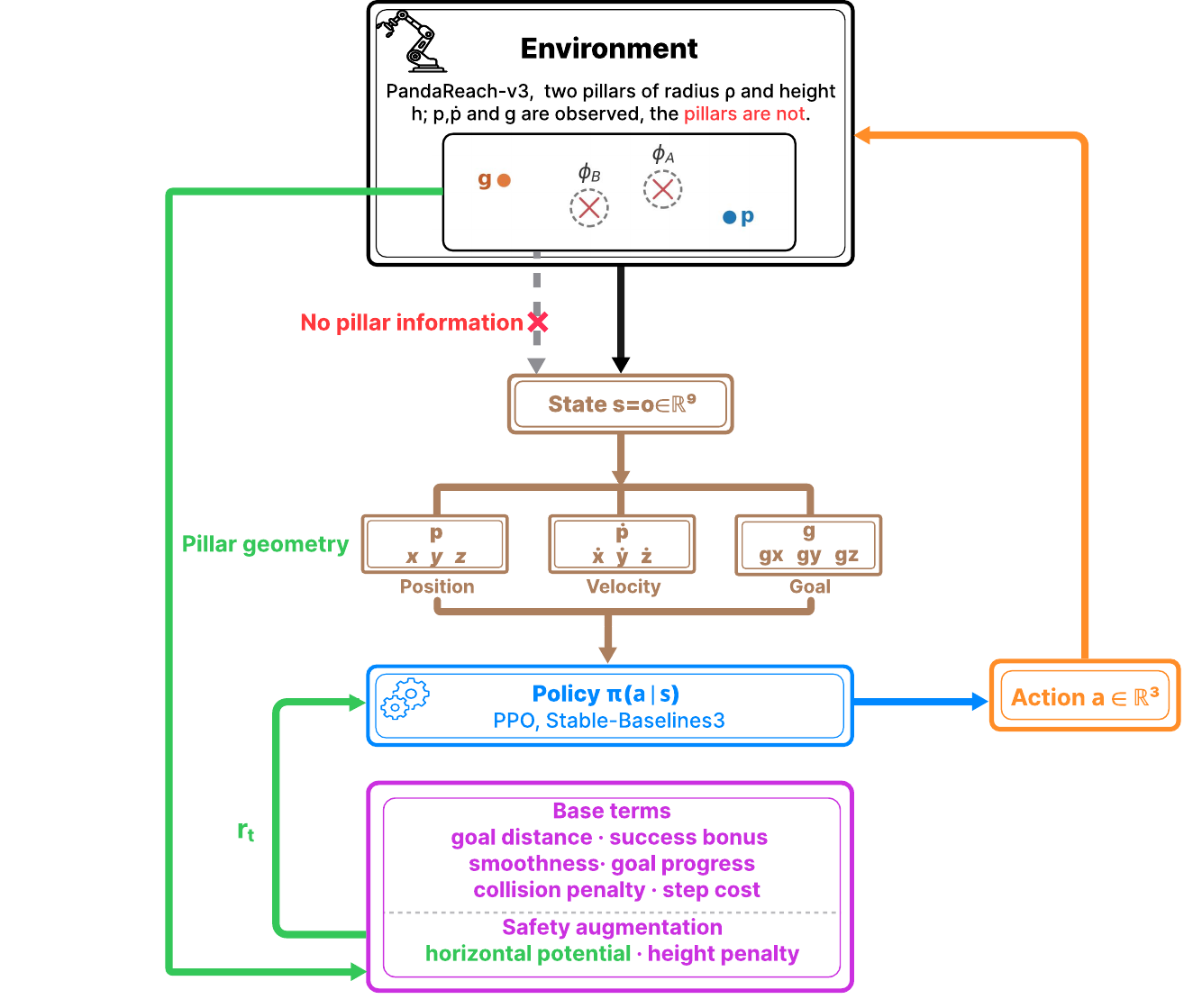}
\caption{Hidden-observation training loop. The policy maps the $9$-dimensional
state (end-effector position and velocity, goal) to a $3$-dimensional action; the
pillars are absent from the state. Obstacle information reaches the policy only
through the reward: the collision penalty and the safety augmentation (horizontal
potential $\Phi$ and height penalty $H$).}\label{fig:overview}
\end{figure}

Figure~\ref{fig:overview} summarizes the resulting loop. At every step the
environment returns the state $\mathbf{s} = \mathbf{o} \in \mathbb{R}^{9}$, which
consists of the end-effector position $\mathbf{p}$, its velocity
$\dot{\mathbf{p}}$, and the goal $\mathbf{g}$; the pillars are present in the scene
but not in $\mathbf{s}$. The policy $\pi(\mathbf{a} \mid \mathbf{s})$, trained with
PPO~\cite{schulman2017}, outputs the action $\mathbf{a} \in \mathbb{R}^{3}$, and the
environment executes it. The reward $r_t$ is computed from the terms listed in the
lower block and returned to the policy. The pillar geometry enters this reward
through the collision penalty and the safety augmentation, that is, the horizontal
potential $\Phi$ and the height penalty $H$ defined in
Section~\ref{sec:method-apf}; $\Phi$ is the only term that penalizes proximity to
a pillar before contact.

\subsection{Problem Formulation}\label{sec:method-mdp}

We formulate size-robust obstacle avoidance in reaching as a finite-horizon
continuous Markov decision process
$\mathcal{M} = (\mathcal{S}, \mathcal{A}, P, r, \gamma)$.
A 7-degree-of-freedom Franka Emika Panda manipulator, simulated in PyBullet
through \emph{panda-gym}~\cite{gallouedec2021} (\texttt{PandaReach-v3}), must guide its end-effector
to a sampled goal while avoiding two cylindrical pillars; the task geometry and
the pillar motion are specified in Section~\ref{sec:exp-setup}.
The state $\mathbf{s} \in \mathcal{S}$ comprises the end-effector position
$\mathbf{p} = (x, y, z) \in \mathbb{R}^{3}$ (m), its velocity
$\dot{\mathbf{p}} \in \mathbb{R}^{3}$ ($\mathrm{m\,s^{-1}}$), and the goal
position $\mathbf{g} \in \mathbb{R}^{3}$ (m); under hidden observation the
obstacle information is withheld (Section~\ref{sec:method-obs}).
The action $\mathbf{a} \in \mathcal{A} = [-1, 1]^{3}$ is a dimensionless
end-effector displacement command, which the environment scales to a maximum
step of $0.05\,\mathrm{m}$ and maps to joint targets through inverse kinematics.
The transition kernel $P$ combines the deterministic robot dynamics with the
stochastic pillar motion of Section~\ref{sec:exp-setup}, so the effective
transition is stochastic.
The reward $r$ is defined in Section~\ref{sec:method-reward}, the discount
factor is $\gamma = 0.99$ (the default of the solver), and an episode terminates when
the end-effector reaches the goal ($\lVert \mathbf{p} - \mathbf{g} \rVert <
0.03\,\mathrm{m}$) or after $100$ steps.
The $N=2$ pillars have radius $\rho$ and height $h$; their placement and the
goal sampling are specified in Section~\ref{sec:exp-setup}.
Strictly, under hidden observation the task is a partially observable Markov decision
process: the policy is a feed-forward network without memory and conditions on the
current $9$-dimensional observation only, so the instantaneous pillar positions,
which drift as a random walk, cannot be inferred from the observation history.
The nominal pillar centers, in contrast, are identical in every episode and are
therefore available to the policy implicitly, through its learned weights; the
quantities that stay hidden are the drift and the radius.

\subsection{Observation Model}\label{sec:method-obs}

\begin{figure}[!t]
\centering
\includegraphics[width=\linewidth]{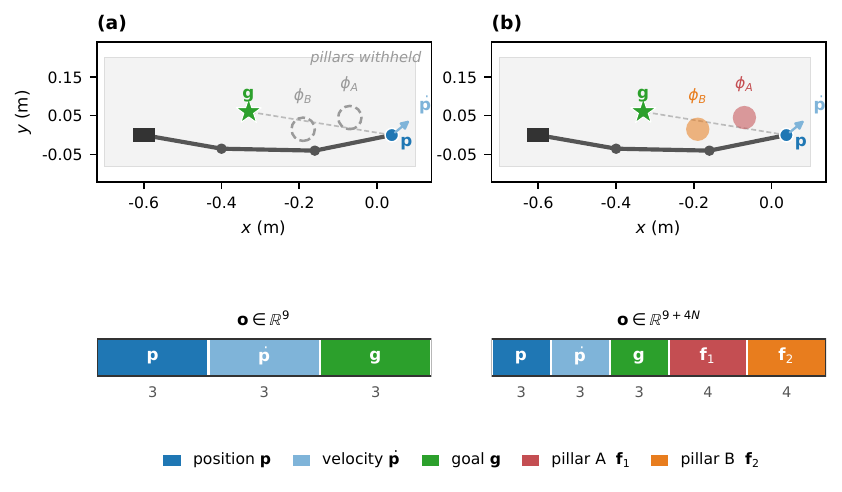}
\caption{Observation model. Top: workspace (top view); the dark polyline
denotes the schematic manipulator from its base to the end-effector at
$\mathbf{p}$. Bottom: observation vector received by the policy, with segment
widths equal to the corresponding dimensions. (a)~Hidden observation
($\mathbb{R}^{9}$): the policy receives $\mathbf{p}$, $\dot{\mathbf{p}}$ and
$\mathbf{g}$; the pillars are withheld. (b)~Full observation
($\mathbb{R}^{9+4N}$): the per-pillar features
$\mathbf{f}_i = [c_{i,x}-x,\ c_{i,y}-y,\ d_{c,i},\ h-z]$ are appended.}\label{fig:obs}
\end{figure}

The environment exposes a dictionary observation composed of three entries: the
end-effector state, the desired goal, and the achieved goal. The end-effector
state is
$\mathbf{o}_{\mathrm{ee}} = [\,\mathbf{p}^{\top},\ \dot{\mathbf{p}}^{\top}\,]^{\top} \in \mathbb{R}^{6}$,
and the desired goal is $\mathbf{g} \in \mathbb{R}^{3}$; the achieved goal
equals the current end-effector position $\mathbf{p}$ and is therefore redundant
with the state. Under the default \emph{hidden observation}, the policy receives
these entries and no obstacle information,
\begin{equation}
\mathbf{o} = \big[\,\mathbf{o}_{\mathrm{ee}}^{\top},\ \mathbf{g}^{\top}\,\big]^{\top} \in \mathbb{R}^{9},
\label{eq:obs-hidden}
\end{equation}
and does not observe the instantaneous location, motion, or size of either pillar. Under
\emph{full observation}, the observation is augmented with the obstacle features
of each pillar,
\begin{equation}
\mathbf{o} = \big[\,\mathbf{o}_{\mathrm{ee}}^{\top},\ \mathbf{g}^{\top},\ \mathbf{f}_{1}^{\top},\ \dots,\ \mathbf{f}_{N}^{\top}\,\big]^{\top} \in \mathbb{R}^{9 + 4N},
\label{eq:obs-full}
\end{equation}
where the feature vector of pillar $i$ is
\begin{equation}
\mathbf{f}_{i} = \big[\,c_{i,x} - x,\ c_{i,y} - y,\ d_{c,i},\ h - z\,\big],
\label{eq:obs-features}
\end{equation}
with $c_{i,x} - x$ and $c_{i,y} - y$ the horizontal offsets from the
end-effector to the center of pillar $i$, $d_{c,i}$ the horizontal clearance of
Equation~\eqref{eq:clearance}, and $h - z$ the vertical distance to the pillar
top. For $N = 2$ pillars, full observation appends $4N = 8$ dimensions.
Figure~\ref{fig:obs} illustrates the two modes.

\subsection{Reward Design}\label{sec:method-reward}

The reward augments the native reaching objective of \texttt{PandaReach-v3}
with a set of auxiliary terms, each introduced below and assembled into the
complete reward at the end of this section. Throughout,
$d_t = \lVert \mathbf{p}_t - \mathbf{g} \rVert$ denotes the distance between the
end-effector and the goal after step $t$, and $\delta = 0.03\,\mathrm{m}$ is the
success threshold.

\subsubsection{Reaching reward}\label{sec:reward-reaching}

The base reward of \texttt{PandaReach-v3} is the negative goal distance $-d_t$,
which drives the end-effector toward the goal at every step.
A one-time success bonus $\lambda_b\,\mathbf{1}[\,d_t < \delta\,]$ is awarded at
the step that terminates the episode at the goal. It makes successful arrival
explicit and offsets the cost of detours around the pillars.

\subsubsection{Goal-progress reward}\label{sec:reward-progress}

The goal-progress term
$\lambda_p\,\max\left(0,\, d_{t-1} - d_t\right)$
rewards any reduction of the goal distance between consecutive steps and adds no
extra penalty when the distance increases, so it remains compatible with detours
that temporarily move away from the goal.

\subsubsection{Smoothness penalty}\label{sec:reward-smoothness}

The smoothness term, weighted by $\lambda_\sigma$, penalizes the squared change in
action magnitude,
\begin{equation}
\sigma_t = \left( \lVert \mathbf{a}_t \rVert - \lVert \mathbf{a}_{t-1} \rVert \right)^{2},
\label{eq:smoothness}
\end{equation}
rather than the change in action direction.
It suppresses abrupt changes in commanded speed while leaving the policy free to
steer around the pillars, since a change of direction at constant magnitude
carries no penalty.

\subsubsection{Artificial potential field}\label{sec:method-apf}

\begin{figure}[!t]
\centering
\includegraphics[width=\linewidth]{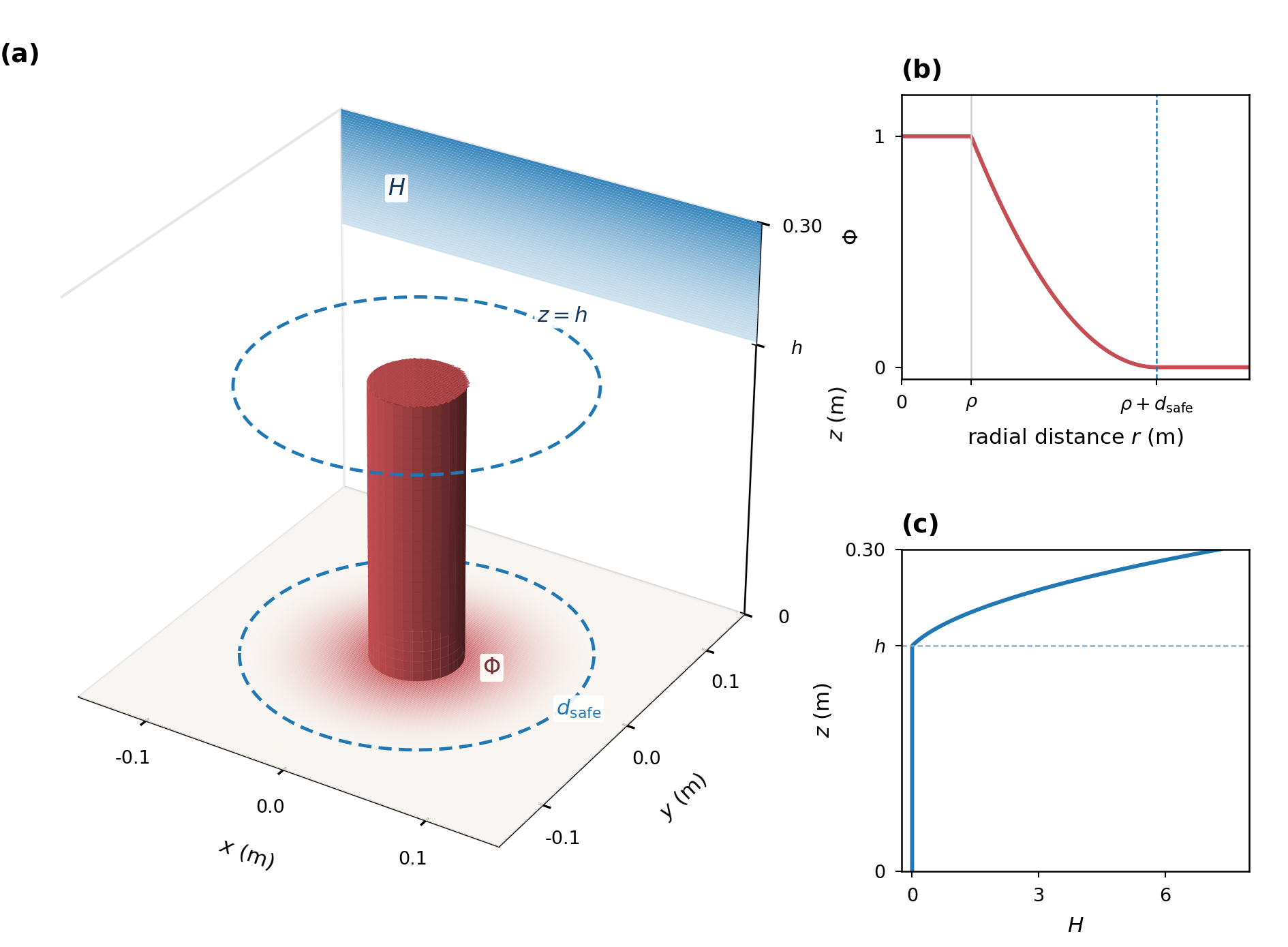}
\caption{Artificial potential field around one pillar. (a)~The horizontal potential
$\Phi$ (red shading) acts within the height band $0\le z\le h$, and the height penalty
$H$ (blue shading) acts above the pillar top; the dashed circle on the floor marks
the outer boundary $r=\rho+d_{\mathrm{safe}}$ of $\Phi$. (b)~$\Phi$ as a function of the horizontal distance $r$ from the pillar
center: it equals one inside the pillar and decreases quadratically to zero at
$r=\rho+d_{\mathrm{safe}}$. (c)~$H$ as a function of the height $z$.}\label{fig:apf}
\end{figure}

The safety prior is injected through two terms that penalize proximity to the
pillars. For pillar $i$, the horizontal clearance
\begin{equation}
d_{c,i}(\mathbf{p}) = \lVert \mathbf{p}_{xy} - \mathbf{c}_{i,xy} \rVert - \rho
\label{eq:clearance}
\end{equation}
is the signed distance between the end-effector and the pillar surface in the
horizontal plane, where $\mathbf{p}_{xy}$ and $\mathbf{c}_{i,xy}$ are the
horizontal positions of the end-effector and of the pillar center, and $\rho$ is
the pillar radius.
The horizontal potential, weighted by $\lambda_s$, is
\begin{equation}
\Phi(\mathbf{p}) = \sum_{i=1}^{N}
\begin{cases}
\left( \dfrac{d_{\mathrm{safe}} - \max(d_{c,i},\, 0)}{d_{\mathrm{safe}}} \right)^{2},
& 0 \le z \le h \ \text{and}\ d_{c,i} < d_{\mathrm{safe}},\\[8pt]
0, & \text{otherwise},
\end{cases}
\label{eq:apf}
\end{equation}
with $d_{\mathrm{safe}} = 0.08\,\mathrm{m}$ and $N = 2$.
It is active only within the pillar's height band ($0 \le z \le h$) and only
when the clearance is smaller than the safety margin.
It penalizes not only penetration ($d_{c,i} < 0$) but also an annular region of
width $d_{\mathrm{safe}}$ outside the pillar surface
($0 \le d_{c,i} < d_{\mathrm{safe}}$), with a quadratic profile that vanishes at
the boundary.
The term therefore favors a positive clearance from the surface over skimming it,
at the cost of a longer path (Section~\ref{sec:exp-margin}).
The potential is defined relative to the pillar surface, but the policy is trained at
a single radius and does not observe that radius. The learned clearance is therefore a fixed
buffer around the surface at the training radius, not a function of the radius at test
time (Section~\ref{sec:exp-margin}).
Figure~\ref{fig:apf} shows the horizontal potential and the height penalty defined
below.

The height penalty encodes the prior that obstacles are to be avoided laterally
rather than by traversing above them. It is a soft ceiling guard, not a hard
constraint:
\begin{equation}
H(\mathbf{p}) = \frac{\bar{z}}{0.04} + \left( \frac{\bar{z}}{0.04} \right)^{2},
\qquad \bar{z} = \max(0,\, z - h),
\label{eq:height}
\end{equation}
weighted by $\lambda_h$. It is zero below the pillar top and grows linearly and
quadratically with the overshoot above it, so it raises the cost of the over-top
route without forbidding it. Whether the over-top route is avoided in practice is
therefore an empirical question, not an assumption, and it is examined together
with the horizontal term in Sections~\ref{sec:exp-setup} and~\ref{sec:exp-xy}.

\subsubsection{Collision penalty and step cost}\label{sec:reward-collision}

The collision indicator is
\begin{equation}
c_t = \bigvee_{i=1}^{N} \left( d_{c,i} < 0 \ \land\ 0 \le z \le h \right),
\label{eq:collision}
\end{equation}
that is, the end-effector collides when its horizontal position lies inside a
pillar while its height lies within the pillar's height band.
A collision is non-terminal: the pillars carry no collision body, so the
end-effector may pass through a pillar and still reach the goal, and each step
inside a pillar incurs the penalty $\lambda_c$ without ending the episode.
Collisions are non-terminal because in preliminary runs with terminal collisions
the policy without the APF term collapsed to a $100\%$ collision rate during
training, so the comparison would reflect a failure to learn rather than the
absence of the safety prior.
The constant step cost $\lambda_t$ makes long detours expensive, so the
learned policy must trade path length against safety.

Combining the terms above, the agent receives the scalar reward
\begin{equation}
\begin{aligned}
r_t ={} & -d_t + \lambda_b\,\mathbf{1}[\,d_t < \delta\,] + \lambda_p\,\max\!\left(0,\, d_{t-1} - d_t\right) \\
        & - \lambda_\sigma\,\sigma_t - \lambda_s\,\Phi(\mathbf{p}_t) - \lambda_h\,H(\mathbf{p}_t)
          - \lambda_c\,\mathbf{1}[\,c_t\,] - \lambda_t,
\end{aligned}
\label{eq:reward}
\end{equation}
with the default coefficients of Table~\ref{tab:reward}.

\begin{table}[htbp]
\centering
\caption{Reward terms and their default coefficients in
Equation~\eqref{eq:reward}.}\label{tab:reward}
\begin{tabular}{llc}
\toprule
Term & Meaning & Default value\\
\midrule
$-d_t$ & dense distance to goal & $1$ (native)\\
$\lambda_b\,\mathbf{1}[\,d_t < \delta\,]$ & success bonus & $5.0$\\
$\lambda_p\,\max(0,\, d_{t-1} - d_t)$ & goal-progress reward & $1.0$\\
$-\lambda_\sigma\,\sigma_t$ & action-magnitude smoothness & $0.03$\\
$-\lambda_s\,\Phi(\mathbf{p}_t)$ & horizontal APF potential & $1.5$\\
$-\lambda_h\,H(\mathbf{p}_t)$ & over-the-top height penalty & $2.0$\\
$-\lambda_c\,\mathbf{1}[\,c_t\,]$ & collision penalty & $0.2$\\
$-\lambda_t$ & step cost & $0.05$\\
\bottomrule
\end{tabular}
\end{table}

The assembly of these terms into the per-step reward is given as pseudocode in
Algorithm~\ref{alg:apf-reward} (\ref{app:algorithm}).

\subsection{Reward Configurations}\label{sec:method-groups}

To attribute the observed behavior to individual reward terms, the full reward
(\ours) is compared with four configurations, each obtained by removing
one or more terms from Equation~\eqref{eq:reward} (Table~\ref{tab:groups}).
\noapf removes both the horizontal potential and the height penalty; \noprog removes the goal-progress reward;
\vanilla removes all auxiliary terms and retains only the dense distance,
the collision penalty, and the step cost; and \noxy removes only the
horizontal potential while retaining the height penalty.
The last control isolates the horizontal potential as the single variable of
interest, which separates its contribution from that of the over-the-top penalty.

\begin{table}[htbp]
\centering
\caption{Reward configurations obtained by removing terms from
Equation~\eqref{eq:reward}.}\label{tab:groups}
\begin{tabular}{@{}ll@{}}
\toprule
Configuration & Removed terms\\
\midrule
\ours & none\\
\noapf & $\lambda_s\,\Phi(\mathbf{p}_t) + \lambda_h\,H(\mathbf{p}_t)$\\
\noprog & $\lambda_p\,\max(0,\, d_{t-1} - d_t)$\\
\vanilla & all auxiliary terms\\
\noxy & $\lambda_s\,\Phi(\mathbf{p}_t)$\\
\bottomrule
\end{tabular}
\end{table}


\section{Experimental Evaluation}\label{sec:experiments}

The experiments are designed to test three hypotheses.
\emph{H1 (robustness).}
The APF reward term confers robustness to out-of-distribution obstacle sizes:
policies trained with the potential maintain safe operation throughout the
evaluation range, whereas policies trained without it deteriorate markedly once the
obstacle radius exceeds its training value (Section~\ref{sec:exp-main}).
\emph{H2 (mechanism).}
This robustness originates from a positive safety margin: the horizontal APF term
moves the learned optimum from a near-zero clearance to a clearance of several
centimeters from the obstacle surface at the training radius, and the resulting
improvement in safe success rate is statistically significant at all
out-of-distribution radii, whereas removing only the horizontal term eliminates it
(Sections~\ref{sec:exp-margin}--\ref{sec:exp-xy}).
\emph{H3 (generality).}
The effect is not specific to a single configuration: it persists under full
observation and across three reinforcement learning solvers
(Section~\ref{sec:exp-generality}), and the learned policy, which does not observe the
obstacles, attains a safe success rate comparable to that of a classical APF planner
that reads the pillar positions (Section~\ref{sec:exp-classic}).
The remainder of this section describes the experimental setup
(Section~\ref{sec:exp-setup}) and then presents the evidence for each hypothesis.

\subsection{Experimental Setup}\label{sec:exp-setup}

\begin{figure}[!t]
\centering
\includegraphics[width=0.9\linewidth]{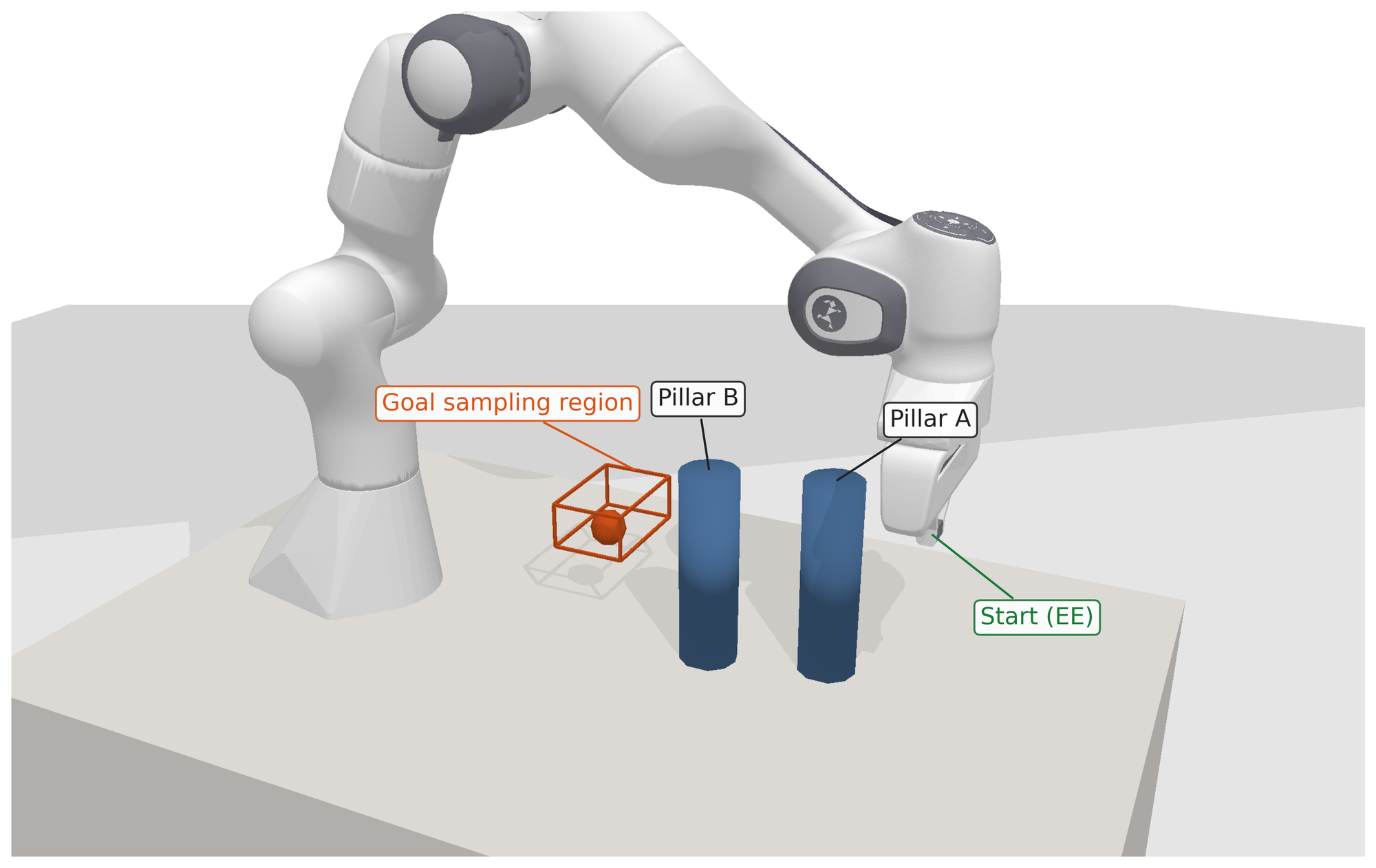}
\caption{Rendered experimental scene (PyBullet, \emph{panda-gym}
\texttt{PandaReach-v3}). The end-effector starts at the position marked ``Start
(EE)'' and is required to reach a goal sampled inside the orange box (the orange sphere is one
example goal), passing the two staggered pillars.}\label{fig:scene}
\end{figure}

\textbf{Task and environment.}
We evaluate on a simulated 7-DoF Panda reaching task (\texttt{PandaReach-v3} from
\emph{panda-gym}).
The policy controls the end-effector (EE) to reach a sampled goal while avoiding two
cylindrical obstacles (pillars).
The two pillars $\phi_A$ and $\phi_B$ have nominal centers at
$(x,y)=(-0.07,0.045)\,\mathrm{m}$ and $(-0.19,0.015)\,\mathrm{m}$ and a height of
$0.21\,\mathrm{m}$ (Figure~\ref{fig:scene}).

\begin{figure}[!t]
\centering
\includegraphics[width=\linewidth]{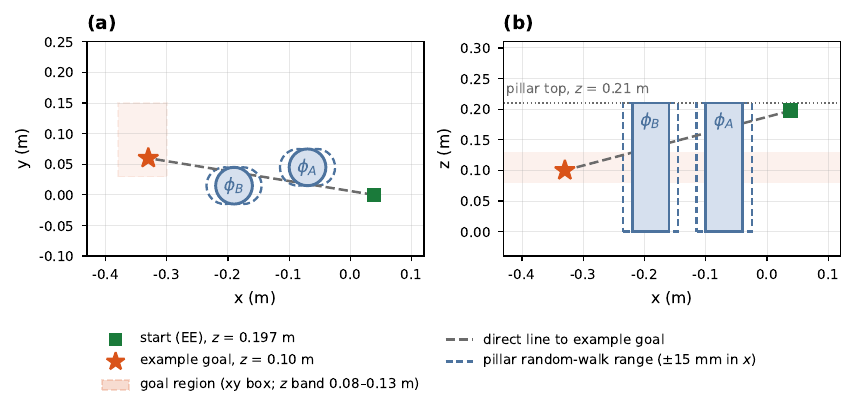}
\caption{Task geometry at the training radius ($\rho=0.03\,\mathrm{m}$).
(a)~Top view: the direct line from the start to the example goal passes within $\rho$
of both pillar centers.
(b)~Side view (projection onto the $xz$ plane): the pillars are $0.21\,\mathrm{m}$ high,
and all goals are sampled below the pillar top.
Dashed outlines show the range of pillar-center motion ($\pm15\,\mathrm{mm}$ along $x$);
the shaded regions indicate the goal sampling area in (a) and the goal height range
in (b).}\label{fig:setup}
\end{figure}

The pillars are staggered diagonally such that, for every sampled goal, the straight line from
the start pose intersects at least one pillar (pillar~$\phi_A$ for $82\%$ of the
sampled goals, pillar~$\phi_B$ for $38\%$, and both for $20\%$, evaluated at the
nominal centers with $\rho=0.03\,\mathrm{m}$ over the goal-sampling distribution); the EE
must therefore detour around the pillars rather than move straight to the goal
(Figure~\ref{fig:setup}).

\begin{figure}[!t]
\centering
\includegraphics[width=\linewidth]{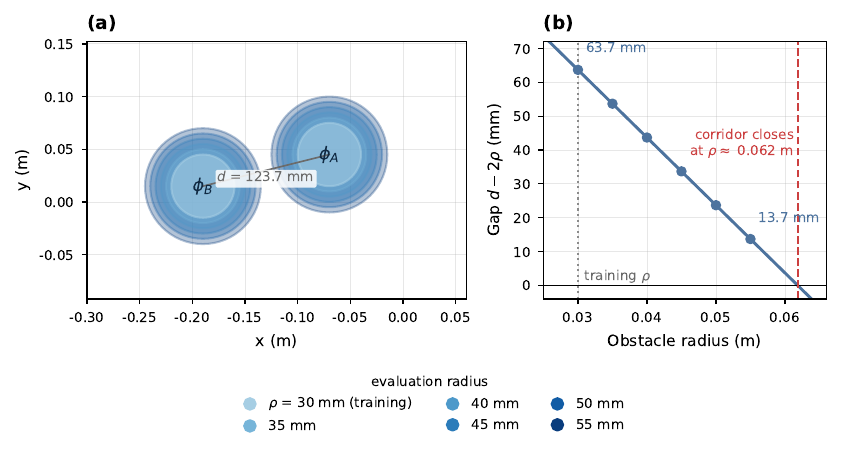}
\caption{Evaluation radius sweep. (a)~Pillar outlines for the six evaluation radii
(top view, nominal pillar centers). (b)~Gap between the two pillar surfaces as a
function of $\rho$ at the nominal centers (markers: the evaluation radii); the corridor
closes at $\rho\approx0.062\,\mathrm{m}$.}\label{fig:sweep}
\end{figure}

Each pillar center performs an independent bounded random walk along the $x$-axis:
at every step, it moves by an increment drawn uniformly from $[-5,5]\,\mathrm{mm}$, and
it is reflected at $\pm15\,\mathrm{mm}$ from its nominal position.
The resulting motion is unpredictable, so a policy cannot rely on exact pillar
positions. The nominal centers, however, are identical in every episode of
training and evaluation (a fixed workcell), so the policy can learn the nominal
layout from experience; only the drift and the radius are unobserved.
The training obstacle radius is $\rho=0.03\,\mathrm{m}$; the evaluation covers six
radii, $\rho\in\{0.03,0.035,0.04,0.045,0.05,0.055\}\,\mathrm{m}$, to test robustness
to out-of-distribution obstacle sizes (Figure~\ref{fig:sweep}).
The upper bound of the sweep is structural rather than arbitrary: the nominal pillar
centers are $0.124\,\mathrm{m}$ apart, so the corridor between the pillar surfaces
narrows from $63.7\,\mathrm{mm}$ at $\rho=0.03\,\mathrm{m}$ to $13.7\,\mathrm{mm}$ at
$\rho=0.055\,\mathrm{m}$ and closes at $\rho\approx0.062\,\mathrm{m}$.
Because the pillars drift by up to $\pm15\,\mathrm{mm}$, the instantaneous corridor
varies around these nominal values: the center distance can fall to $0.095\,\mathrm{m}$,
so the corridor can close transiently for $\rho>0.047\,\mathrm{m}$.

\textbf{Over-the-top traversal.}
The pillars do not reach the workspace ceiling, so an end-effector trajectory above
$z=0.21\,\mathrm{m}$ passes them without collision. No trained policy exploits this route:
the over-top rate is $0\%$ for every policy at every radius, including \noapf and
\vanilla, whose rewards contain no ceiling penalty, and a search over scripted
over-top trajectories indicates that the low-altitude route attains the higher return
under this reward (\ref{app:overtop}).

\textbf{Partial observability.}
Unless stated otherwise, policies are trained and evaluated under
\emph{hidden observation}: the observation retains only the 6-dimensional
end-effector state (position and velocity) and the goal, while the per-pillar obstacle
features $\mathbf{f}_i$ (Equation~\eqref{eq:obs-features}) are withheld.
The policy therefore does not observe the instantaneous position or the size of the
obstacles, and the reward is the only channel through which information about the
obstacle size reaches it (Section~\ref{sec:method-reward}).

\textbf{Training protocol.}
The default solver is proximal policy optimization (PPO~\cite{schulman2017},
Stable-Baselines3 \texttt{MultiInputPolicy}, entropy coefficient $0.01$), trained for
$1.5\times10^5$ steps; a model checkpoint is evaluated every $1.5\times10^4$ steps on
$20$ goals at the training radius.
Collisions are non-terminal (Section~\ref{sec:reward-collision}): each step spent
inside a pillar incurs a penalty $\lambda_c = 0.2$, and the episode continues.
We adopt this design because, in preliminary runs with terminal collisions, the no-APF
policy collapsed to a $100\%$ collision rate during training, and the comparison would
then reflect a failure to learn rather than the safety margin.
A constant step cost $\lambda_t = 0.05$ penalizes long trajectories, so a policy must
trade path length against safety.

\textbf{Implementation details.}
The task is simulated with PyBullet through \emph{panda-gym}
3.0.7~\cite{gallouedec2021} (Gymnasium 1.3.0);
an episode ends when the EE comes within $0.03\,\mathrm{m}$ of the goal or after
$100$ steps.
The algorithms are taken from Stable-Baselines3
2.9.0~\cite{raffin2021} (PyTorch 2.14.0, Python 3.10)
with default hyperparameters unless stated otherwise; the settings of the environment, the solvers, and the classical planner are listed in \ref{app:implementation}.
All experiments run on a laptop with an Intel Core i5-11400H CPU and $16$\,GB RAM
under Windows~11, without a GPU.

\textbf{Metrics.}
Each episode ends in one of four mutually exclusive outcomes, determined by whether
the EE reaches the goal ($S$) and whether it collides with a pillar at any step ($C$):
safe reach ($S\land\neg C$), through-collision reach ($S\land C$), safe timeout
($\neg S\land\neg C$), and collision timeout ($\neg S\land C$).
We use the \emph{safe success rate} as the primary metric,
\begin{equation}
P(S\land\neg C)=P(S)-P(S\land C),
\end{equation}
rather than the raw success rate $P(S)$. Because collisions are non-terminal, the EE
can reach the goal \emph{through} a pillar, so $P(S)$ stays near $1$ even when a policy
collides in $80\%$ of episodes, and $P(S)$ is therefore uninformative about safety.
The collision-timeout outcome ($\neg S\land C$) is rare: across all PPO groups the
collision rate $P(C)=P(S\land C)+P(\neg S\land C)$ exceeds the through-collision
rate $P(S\land C)$ by at most $2.7$ percentage points (seed-averaged), and by at most
$1.0$ percentage point for \ours and \noapf in particular.
Hence $P(S\land C)\approx P(C)$, and the safe success rate is well approximated by
$P(S)-P(C)$.
Because $P(S)$ is itself below one (e.g., $94.3\%$ for \ours), the safe success
rate is not simply $1-P(C)$; we therefore report both the safe success rate and the
collision rate throughout.

\textbf{Statistical treatment.}
We report $95\%$ confidence intervals ($t$-distribution, $n-1$ degrees of freedom)
and two-sided Welch $t$-tests (unequal variances) on the safe success rate; seed
counts are given in each table caption. Standard deviations are sample standard
deviations ($n-1$). Because the safe success rate is bounded and its distribution
across seeds is not normal at the extremes, the principal comparisons are checked with
the rank-based Mann--Whitney $U$ test. The abbreviation n.s.\ denotes a difference
that is not significant at the $0.05$ level.

\textbf{Reward configurations and baselines.}
The full reward (\ours) and the four reward configurations of
Section~\ref{sec:method-groups} are compared in the main sweep
(Section~\ref{sec:exp-main}); Section~\ref{sec:exp-xy} isolates the horizontal
APF term, and Section~\ref{sec:exp-collpen} tests a larger collision penalty as an
alternative source of margin. A classical APF planner (Section~\ref{sec:exp-classic}) and, for the
solver study, SAC and TD3 (Section~\ref{sec:exp-solver}) serve as external
references.

\subsection{Size Robustness of the Safe Success Rate}\label{sec:exp-main}

Table~\ref{tab:main} reports the safe success rate under hidden observation as the
obstacle radius increases beyond its training value.

\begin{figure}[!t]
\centering
\includegraphics[width=\linewidth]{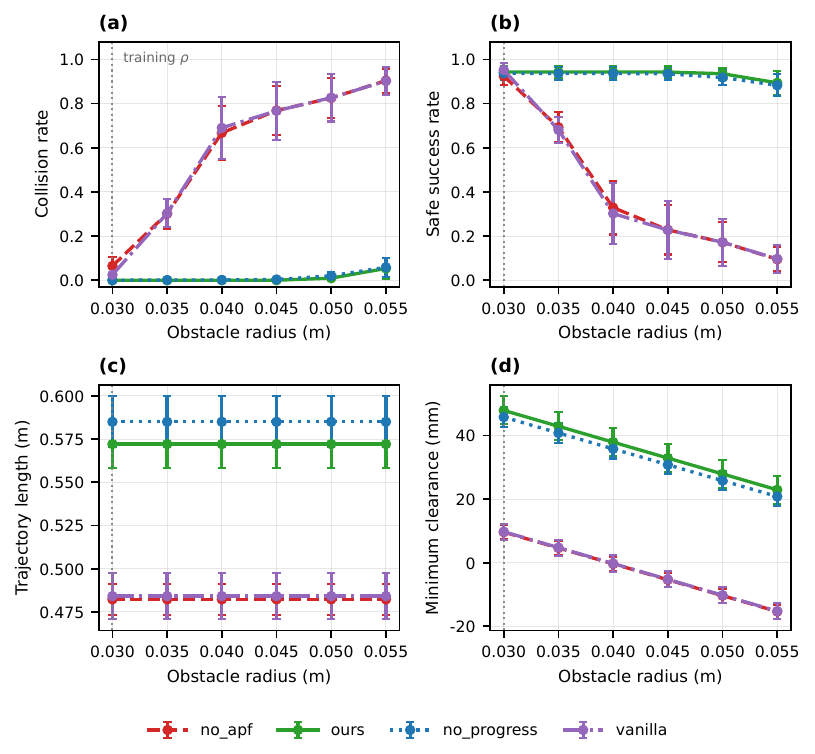}
\caption{(a)~Collision rate, (b)~safe success rate, (c)~trajectory length, and
(d)~minimum clearance vs.\ obstacle radius under hidden observation ($n=30$ seeds per
group; error bars denote $95\%$ confidence intervals).}\label{fig:main}
\end{figure}

Two observations follow from Table~\ref{tab:main}. First, the APF groups
(\ours and \noprog) maintain a safe success rate above $88\%$ over
the whole radius range, whereas the groups without the APF term (\noapf and
\vanilla) deteriorate sharply once $\rho$ exceeds the training value: at
$\rho=0.035\,\mathrm{m}$ their safe success rate has already fallen to $69.3\%$ and
$68.1\%$, and by $\rho=0.055\,\mathrm{m}$ it reaches $9.5\%$ and $9.7\%$.
Their collision rates increase correspondingly from $30.1\%$ and $30.3\%$ at
$\rho=0.035\,\mathrm{m}$ to $90.4\%$ and $90.2\%$ at $\rho=0.055\,\mathrm{m}$, while
the APF groups remain below $6\%$ collision throughout. Second,
\noprog closely matches \ours, so the robustness does not depend on
the goal-progress reward.
Figure~\ref{fig:main} plots the safe success rate, collision rate, trajectory length,
and minimum clearance against the obstacle radius.
\begin{table}[htbp]
\centering
\caption{Safe success rate (\%) under hidden observation as a function of obstacle
radius. Mean over $n=30$ seeds per group, each evaluated on $50$ goals.}\label{tab:main}
\begin{tabular}{lcccccc}
\toprule
Group & \multicolumn{6}{c}{Obstacle radius $\rho$ (m)}\\
\cmidrule(lr){2-7}
& 0.030 & 0.035 & 0.040 & 0.045 & 0.050 & 0.055\\
\midrule
\ours & 94.3 & 94.3 & 94.3 & 94.3 & 93.5 & 89.4\\
\noapf   & 92.5 & 69.3 & 32.9 & 22.8 & 17.2 & 9.5\\
\noprog & 93.7 & 93.7 & 93.7 & 93.5 & 91.8 & 88.3\\
\vanilla & 95.3 & 68.1 & 30.2 & 22.7 & 17.2 & 9.7\\
\noxy & 83.6 & 64.6 & 39.8 & 27.7 & 16.6 & 7.5\\
\bottomrule
\end{tabular}
\end{table}

\subsection{Mechanism: Positive Safety Margin}\label{sec:exp-margin}

To identify \emph{why} the APF groups generalize, we quantify the minimum horizontal
clearance $d_c$ along each trajectory, that is, the smallest value of $d_{c,i}$
(Equation~\eqref{eq:clearance}) over both pillars and over all steps at which the EE lies
within the pillar height band $0 \le z \le h$.
Table~\ref{tab:margin} reports this clearance at the training radius.
\begin{table}[htbp]
\centering
\caption{Minimum horizontal clearance and inferred center distance at the training
radius $\rho=0.03\,\mathrm{m}$ (mean $\pm$ sample standard deviation over $n=30$ seeds
per group).}\label{tab:margin}
\begin{tabular}{lccc}
\toprule
Group & Clearance $d_c$ (mm) & Center distance $d_c+\rho$ (mm)\\
\midrule
\ours & $47.9 \pm 11.8$ & $77.9$\\
\noapf     & $9.6 \pm 5.9$   & $39.6$\\
\noprog& $45.8 \pm 8.1$  & $75.8$\\
\vanilla & $9.7 \pm 6.8$   & $39.7$\\
\noxy & $8.4 \pm 8.0$   & $38.4$\\
\bottomrule
\end{tabular}
\end{table}

The APF groups learn a minimum clearance of roughly $46$--$48\,\mathrm{mm}$
(center distance $\approx76$--$78\,\mathrm{mm}$), that is, they maintain the EE about
$46$--$48\,\mathrm{mm}$ \emph{outside} the pillar surface. By contrast, the groups without
the horizontal APF term learn only $\approx8$--$10\,\mathrm{mm}$ of clearance
(center distance $\approx38$--$40\,\mathrm{mm}$), passing close to the
pillar surface and leaving almost no margin. The clearance ratio is approximately
$5\times$, and the center-distance ratio approximately $2\times$.

\begin{figure}[!t]
\centering
\includegraphics[width=0.65\linewidth]{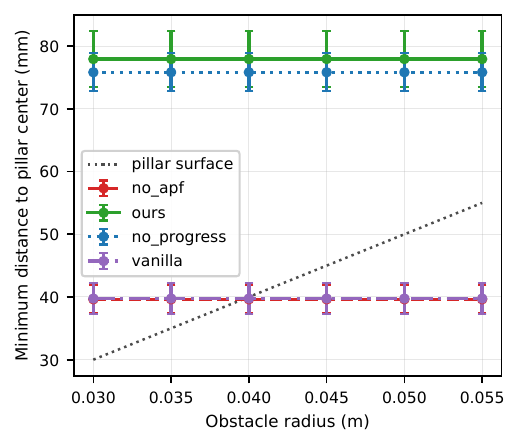}
\caption{Minimum distance from the pillar center vs.\ obstacle radius ($n=30$ seeds
per group; error bars denote $95\%$ confidence intervals). The dotted diagonal marks the
pillar surface, i.e., zero clearance; the curves of \noapf and \vanilla
coincide.}\label{fig:margin}
\end{figure}

The dependence of the clearance on the obstacle radius corroborates the mechanism
(Figure~\ref{fig:margin}). The policy does not observe the pillars, so the executed
trajectory depends on the goal alone and is identical at every obstacle radius.
Consequently, the minimum clearance of \ours
decreases in $\rho$ with a slope of exactly $-1$ (the center distance $d_c+\rho$
remains constant at $77.9\,\mathrm{mm}$), while the center distance of \noapf
remains constant at $39.6\,\mathrm{mm}$. Both policies therefore hold a fixed absolute
distance from the pillar center at test time, because the radius is unobserved; what the
APF term changes is the size of the buffer that the policy has learned at the training
radius. The buffer of \ours ($47.9\,\mathrm{mm}$) exceeds the maximum radius
increase of the sweep ($25\,\mathrm{mm}$), so its mean minimum clearance remains
positive up to $\rho=0.055\,\mathrm{m}$ ($22.9\,\mathrm{mm}$). The mean minimum clearance
of \noapf ($9.6\,\mathrm{mm}$) becomes negative near $\rho=0.04\,\mathrm{m}$, and
$63\%$ of its seeds have a negative mean minimum clearance at that radius, against
$13\%$ at $\rho=0.035\,\mathrm{m}$, where its safe success rate has already fallen to
$69.3\%$.

\begin{figure}[!t]
\centering
\includegraphics[width=\linewidth]{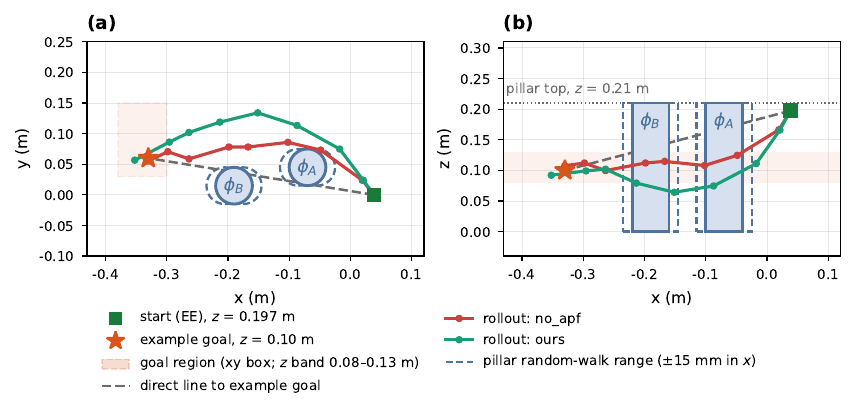}
\caption{Deterministic rollouts of one trained policy per group (PPO, seed~1) toward
the example goal at the training radius $\rho=0.03\,\mathrm{m}$. (a)~Top view.
(b)~Side view (projection onto the $xz$ plane). Pillars are drawn at their nominal
centers.}\label{fig:rollout}
\end{figure}

Figure~\ref{fig:rollout} illustrates the same effect on the example goal.
Both policies pass the pillars below their tops, but the \ours trajectory
swings visibly wider around the pillars than the \noapf trajectory, which is
consistent with the clearances in Table~\ref{tab:margin}.
The figure shows a single seed and serves as an illustration only; the quantitative
evidence is the multi-seed clearances in Table~\ref{tab:margin}.

\subsection{Control: A Larger Collision Penalty}\label{sec:exp-collpen}

\begin{figure}[!t]
\centering
\includegraphics[width=\linewidth]{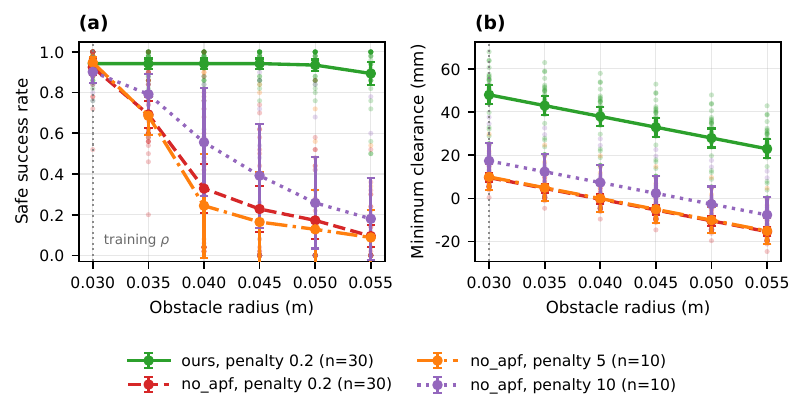}
\caption{Effect of the collision penalty on the \noapf reward (hidden
observation, PPO). (a)~Safe success rate and (b)~minimum clearance vs.\ obstacle radius.
Lines show the mean over seeds ($n=30$ for the default penalty and for \ours,
$n=10$ for penalties $5$ and $10$), error bars denote $95\%$ confidence intervals, and
faint points are the individual seeds.}\label{fig:collpen}
\end{figure}

A natural alternative to the potential is to make contact more expensive: because collisions
are non-terminal and carry a per-step penalty of only $0.2$, passing through a pillar is cheap
under the reward without the potential, and a larger penalty might by itself push the
learned trajectory away from the pillars.
We therefore retrain the \noapf reward (no horizontal potential, no height penalty)
under hidden observation with the collision penalty raised from $0.2$ to $5$ and to $10$
($25\times$ and $50\times$), keeping the training radius, step cost, and all other settings
unchanged ($n=10$ seeds per penalty, PPO). A preliminary run with a penalty of $1$
($n=2$ seeds) produced no margin (clearance $5.5\,\mathrm{mm}$ at the training radius, safe
success rate $0\%$ at $\rho\ge0.045\,\mathrm{m}$) and was not extended.
Figure~\ref{fig:collpen} shows the safe success rate and the minimum clearance across
the sweep for each penalty, and Table~\ref{tab:collpen} compares the results with
\noapf at the default penalty and with \ours.
\begin{table}[htbp]
\centering
\caption{Effect of the collision penalty on the \noapf reward (PPO, hidden
observation). Clearance is the minimum horizontal clearance at the training radius
$\rho=0.03\,\mathrm{m}$ (mean $\pm$ sample standard deviation over seeds); safe success rate
in \%. The last column counts the seeds whose safe success rate at
$\rho=0.055\,\mathrm{m}$ is at least $50\%$.}\label{tab:collpen}
\begin{tabular}{lccccccc}
\toprule
Reward & Penalty & $n$ & Clearance (mm) & \multicolumn{3}{c}{Safe success at $\rho$ (m)} & Seeds $\ge50\%$\\
\cmidrule(lr){5-7}
& & & & 0.030 & 0.040 & 0.055 & at 0.055\\
\midrule
\noapf & $0.2$ & $30$ & $9.6 \pm 5.9$  & 92.5 & 32.9 & 9.5  & $1/30$\\
\noapf & $5$   & $10$ & $9.9 \pm 8.6$  & 94.8 & 24.4 & 8.8  & $1/10$\\
\noapf & $10$  & $10$ & $17.3 \pm 11.4$ & 90.2 & 55.6 & 18.0 & $2/10$\\
\ours    & $0.2$ & $30$ & $47.9 \pm 11.8$ & 94.3 & 94.3 & 89.4 & $30/30$\\
\bottomrule
\end{tabular}
\end{table}

Raising the penalty does not reproduce the effect of the potential.
At a penalty of $5$, the clearance ($9.9\,\mathrm{mm}$) and the safe success rates are
indistinguishable from those at the default penalty, and only $1$ of $10$ seeds retains a
safe success rate above $50\%$ at $\rho=0.055\,\mathrm{m}$, the same as the single such seed
among the $30$ default runs.
A penalty of $10$ yields a larger mean clearance ($17.3\,\mathrm{mm}$) and a higher safe
success rate at the intermediate radii ($55.6\%$ at $\rho=0.04\,\mathrm{m}$), and it lowers the
collision rate at the training radius from $6.5\%$ to $1.0\%$, but the benefit is uneven
across seeds: $3$ of $10$ seeds exceed $20\,\mathrm{mm}$ of clearance and $2$ of $10$ stay above
$50\%$ at the maximum tested radius, whereas all $30$ seeds of \ours do.
At $\rho=0.055\,\mathrm{m}$, both penalties remain far below \ours (Welch $p=2.1\times10^{-8}$ and
$1.1\times10^{-5}$; Mann--Whitney $p=2.9\times10^{-6}$ and $7.7\times10^{-6}$ for penalties
$5$ and $10$, respectively).
A larger penalty therefore changes the margin only partially and unreliably, whereas the
potential produces a margin in every seed.
The result is limited to the two penalty values tested; we cannot exclude that a still larger
penalty, or a combination with other changes, would approach the margin of \ours, although
a very large penalty may also overwhelm the goal-reaching signal.

\subsection{Sample Efficiency}\label{sec:exp-efficiency}

\begin{figure}[!t]
\centering
\includegraphics[width=\linewidth]{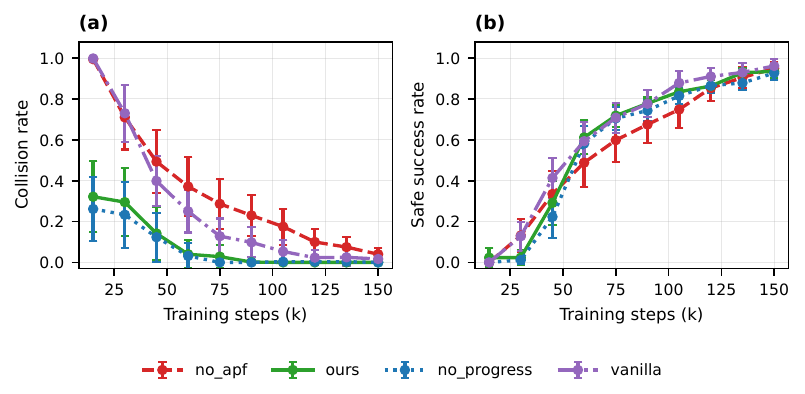}
\caption{Learning curves: (a)~collision rate and (b)~safe success rate vs.\ training
steps ($n=30$ seeds per group, $20$ evaluation goals per checkpoint; error bars denote
$95\%$ confidence intervals). The APF groups
reach zero collision at a median of $1.5\times10^4$ steps, whereas \noapf
reaches it at a median of $1.2\times10^5$ steps among the $21$ converging seeds, and
$9/30$ seeds do not converge within the $1.5\times10^5$-step budget.}\label{fig:curve}
\end{figure}

Beyond the asymptotic robustness gap, the APF term also accelerates the elimination of
collisions.
Figure~\ref{fig:curve} shows the collision rate and safe success rate as a function
of training steps, evaluated on $20$ goals per checkpoint at the training radius (the
radius sweeps use $50$ goals).
For each seed, we take the first checkpoint from which the collision rate stays at zero
through the end of training.
Across the $n=30$ seeds of \ours, the median of this step is $1.5\times10^4$,
which is the first evaluation point ($16$ of the $30$ seeds are collision-free from
there on); all $30$ seeds converge, the latest at $1.05\times10^5$, and the final
collision rate is $0.0\%$.
For \noapf, $21$ of $30$ seeds reach zero collision within the budget, at a
median of $1.2\times10^5$ steps among these seeds, whereas the other $9$ are still above
zero at $1.5\times10^5$ steps (final mean collision rate $4.0\%$).
Counting the $9$ non-converged seeds at the end of the budget, the median over all $30$
seeds is $1.275\times10^5$ steps (Mann--Whitney $U$ test on the convergence step,
$p<10^{-9}$).
The median convergence gap is therefore at least $8\times$, because the first evaluation
point bounds the median of \ours from above and the budget censors that of
\noapf.
\noprog matches \ours (median $1.5\times10^4$, all $30$ seeds
converge), so the speedup does not depend on the progress term.
The speedup concerns collisions only: the safe success rate of \ours and
\noapf does not differ significantly (Welch $p\ge0.05$) at $7$ of the $10$
checkpoints, and \ours reaches the goal less often than \noapf at the end
of training ($93.5\%$ vs.\ $99.0\%$). The goals that \ours fails to reach
concentrate at the far edge of the goal region ($84\%$ reached for $x<-0.355\,\mathrm{m}$
against $98\%$ elsewhere, seeds~1--30), whereas \noapf reaches both parts at
the same rate ($99\%$).

\subsection{Statistical Significance}\label{sec:exp-significance}

Table~\ref{tab:significance} reports $95\%$ confidence intervals and the two-sided
Welch $t$-test between \ours and \noapf on the safe success rate.
\begin{table}[htbp]
\centering
\caption{Safe success rate: $95\%$ confidence interval and two-sided Welch $t$-test
(\ours vs.\ \noapf, $n=30$ each).}\label{tab:significance}
\begin{tabular}{lccccc}
\toprule
 & \multicolumn{2}{c}{\noapf} & \multicolumn{2}{c}{\ours} & \\
\cmidrule(lr){2-3}\cmidrule(lr){4-5}
$\rho$ (m) & Mean (\%) & 95\% CI & Mean (\%) & 95\% CI & $p$\\
\midrule
0.030 & 92.5 & $[88.5,96.5]$ & 94.3 & $[91.7,96.8]$ & $0.44$ (n.s.)\\
0.035 & 69.3 & $[62.5,76.0]$ & 94.3 & $[91.7,96.8]$ & $2.2\times10^{-8}$\\
0.040 & 32.9 & $[20.7,45.0]$ & 94.3 & $[91.7,96.8]$ & $2.0\times10^{-11}$\\
0.045 & 22.8 & $[11.7,33.9]$ & 94.3 & $[91.7,96.8]$ & $4.0\times10^{-14}$\\
0.050 & 17.2 & $[8.1,26.3]$ & 93.5 & $[90.8,96.3]$ & $9.8\times10^{-18}$\\
0.055 & 9.5 & $[4.0,14.9]$ & 89.4 & $[83.9,94.9]$ & $8.0\times10^{-29}$\\
\bottomrule
\end{tabular}
\end{table}

Two points follow from Table~\ref{tab:significance}.
At the training radius ($\rho=0.03\,\mathrm{m}$), the difference is
$+1.8$ percentage points with $p=0.44$, i.e., not significant. This is consistent with
the positive-safety-margin mechanism: the buffer matters only once the obstacle radius
exceeds the training value, so no advantage is expected inside the training
distribution.
At every OOD radius, the advantage is large: from $\rho=0.035\,\mathrm{m}$ onward,
the gap widens from $+25.0$ to $+79.9$ percentage points, with all
$p\le2.2\times10^{-8}$ (four of the five below $10^{-10}$).
The Mann--Whitney $U$ test leads to the same conclusion ($p<10^{-7}$ at every OOD
radius), and all five $p$-values remain significant after a Bonferroni correction
over the five OOD radii.

\subsection{Attribution to the Horizontal Potential}\label{sec:exp-xy}

\begin{figure}[!t]
\centering
\includegraphics[width=\linewidth]{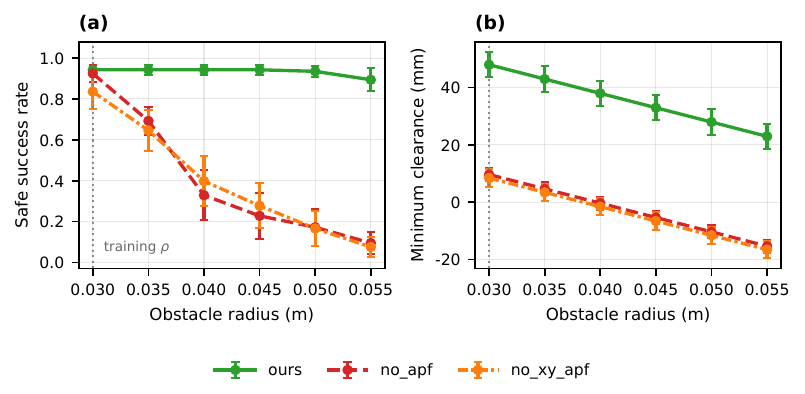}
\caption{Safe success rate (a) and minimum clearance (b) vs.\ obstacle radius for
\ours, \noapf, and \noxy ($n=30$ seeds per group; error bars
denote $95\%$ confidence intervals).}\label{fig:xy}
\end{figure}

The comparisons above attribute the robustness to the APF term as a whole. However,
\noapf removes two reward terms at once, the horizontal APF potential and the
vertical ceiling penalty, which could confound this attribution (Figure~\ref{fig:xy}).
The \noxy control therefore removes only the horizontal APF term and retains
the ceiling penalty. Its safe success rate shows no significant difference from that of
\noapf at any radius (Welch $p$ between $0.065$ and $0.92$), whereas
\ours exceeds \noxy by $+29.7$ to $+81.9$ percentage points at every
OOD radius ($p<1.1\times10^{-6}$) and by $+10.7$ percentage points already at the
training radius ($p=0.021$, not corrected for multiple comparisons). In addition, the
EE never rises above $0.197\,\mathrm{m}$ in any evaluated trajectory
(Section~\ref{sec:exp-setup}), so the ceiling penalty is inactive. The vertical term is therefore not the source of the robustness,
and the positive safety margin is attributable to the horizontal APF potential.

\subsection{Comparison with Classical APF Planning}\label{sec:exp-classic}

\begin{figure}[!t]
\centering
\includegraphics[width=\linewidth]{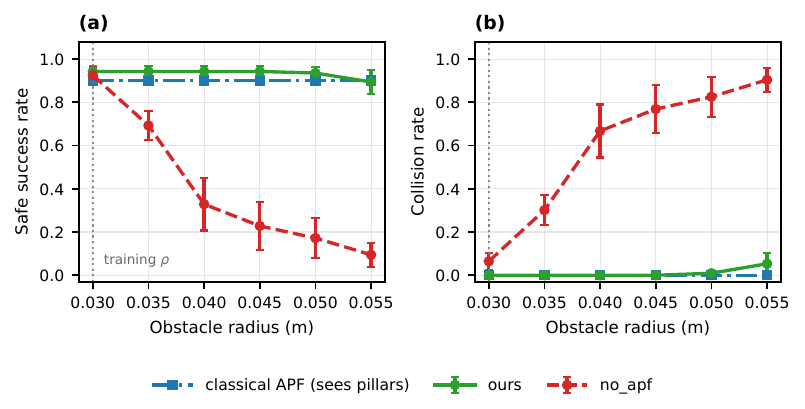}
\caption{(a) Safe success rate and (b) collision rate vs.\ obstacle radius for the
classical APF planner (full pillar state, $50$ goals, deterministic) and the RL policies
(hidden observation, $n=30$ seeds; error bars denote $95\%$ confidence
intervals).}\label{fig:classic}
\end{figure}

The preceding sections attribute the benefit to the horizontal APF term, which raises
the question of how a classical APF controller that reads the obstacles performs under
the same protocol. We therefore implement a Khatib-type
planner~\cite{khatib1986} that computes, at every step, a resultant command from an
attractive term toward the goal and a repulsive term from each pillar.
The attraction is proportional to the goal offset with gain $K_{\mathrm{att}}=20$, the
repulsion has gain $K_{\mathrm{rep}}=5$ within an influence range of $0.08\,\mathrm{m}$
from the pillar surface, the height is driven toward the goal height with a proportional
gain, and the resultant command is saturated at unit norm.
The planner reads the instantaneous pillar centers and the radius, i.e., it operates
with full knowledge of the obstacle state, whereas the RL policies observe nothing about
the pillars.
It runs in the same environment, on the same $50$ goals and six radii, with the same
$100$-step limit and the same collision counting as the RL policies, and its parameters
are fixed over the sweep.
Table~\ref{tab:classic} and Figure~\ref{fig:classic} compare its safe success rate and collision rate with those of the RL policies.
\begin{table}[htbp]
\centering
\caption{Safe success rate (\%) of the classical APF planner (deterministic, $50$
goals, full pillar state) and of the RL policies (hidden observation, mean over $n=30$
seeds).}\label{tab:classic}
\begin{tabular}{lcccccc}
\toprule
 & \multicolumn{6}{c}{Obstacle radius $\rho$ (m)}\\
\cmidrule(lr){2-7}
Method & 0.030 & 0.035 & 0.040 & 0.045 & 0.050 & 0.055\\
\midrule
Classical APF (sees pillars) & 90.0 & 90.0 & 90.0 & 90.0 & 90.0 & 90.0\\
RL \ours  & 94.3 & 94.3 & 94.3 & 94.3 & 93.5 & 89.4\\
RL \noapf & 92.5 & 69.3 & 32.9 & 22.8 & 17.2 & 9.5\\
\bottomrule
\end{tabular}
\end{table}

The classical planner never collides and its safe success rate is $90\%$ at every
radius, which is comparable to that of \ours ($94.3\%$ up to $\rho=0.045\,\mathrm{m}$
and $89.4\%$ at $\rho=0.055\,\mathrm{m}$).
We do not claim that the learned policy outperforms the planner: a $4.3$ percentage-point
difference corresponds to two of the $50$ goals, and the planner is deterministic, so
no confidence interval applies.
The unreached $10\%$ of goals are not caused by the pillars: the same controller with
the repulsive gain set to zero reaches $90\%$ of the goals as well (and collides in all
episodes), and varying $K_{\mathrm{rep}}$ from $0.01$ to $100$ at the training radius
leaves the collision rate at $0\%$ and the safe success rate between $88\%$ and
$90\%$.
The two approaches differ in what they require and in the paths they produce.
The planner adapts to the radius, because it reads it: its minimum clearance stays near
$49\,\mathrm{mm}$ at every $\rho$, at the cost of a longer path that grows from
$1.22\,\mathrm{m}$ to $1.46\,\mathrm{m}$ as $\rho$ increases.
The RL policy needs no obstacle information at run time and executes a much shorter
path ($0.57\,\mathrm{m}$), but it does not adapt: its clearance falls from
$47.9\,\mathrm{mm}$ to $22.9\,\mathrm{mm}$ across the sweep, and its collision rate
becomes non-zero at the largest radii ($0.9\%$ at $\rho=0.05\,\mathrm{m}$ and
$5.3\%$ at $\rho=0.055\,\mathrm{m}$).
The classical planner is therefore a reference that shows what is attainable when the
obstacle state is known; the contribution of the APF reward is to transfer a part of
this safety to a policy that does not observe the obstacles.

\subsection{Generality Across Observation Models and Solvers}\label{sec:exp-generality}

The results so far were obtained with PPO under hidden observation. We now test
whether the effect depends on either choice, first by changing the observation model and
then by changing the learning algorithm.

\subsubsection{Full observation}\label{sec:exp-fullobs}

\begin{figure}[!t]
\centering
\includegraphics[width=0.65\linewidth]{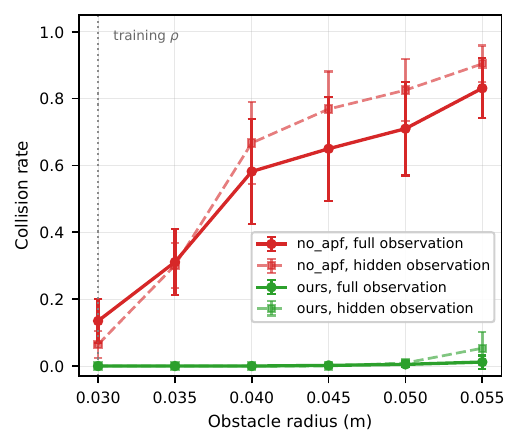}
\caption{Collision rate vs.\ obstacle radius under full observation ($n=20$ seeds)
and hidden observation ($n=30$ seeds). Error bars denote $95\%$ confidence
intervals.}\label{fig:fullobs}
\end{figure}

A possible objection is that the APF advantage is an artifact of the hidden
observation. We therefore retrain \ours and \noapf with
\emph{full observation}, i.e., with the obstacle features $\mathbf{f}_i$
(Equation~\eqref{eq:obs-features}) appended to the state
($n=20$ seeds, same training protocol; each trained policy is evaluated on the same
$50$ goals with a different pillar random-walk realization per goal).
Full observation does not rescue \noapf: its collision rate is $13.5\%$ at the
training radius and $58.2\%$/$83.1\%$ at $\rho=0.04$/$0.055\,\mathrm{m}$, which does not
differ significantly from the hidden-observation values ($6.5\%$, $66.7\%$, $90.4\%$;
Welch $p\ge0.066$ at every radius; Figure~\ref{fig:fullobs}).
\ours maintains a collision rate of $0\%$ up to $\rho=0.04\,\mathrm{m}$ and at most
$1.2\%$ at $\rho=0.055\,\mathrm{m}$, and its collision rate is lower than that of
\noapf at every radius ($p<4.1\times10^{-4}$, Welch; $p<10^{-5}$, Mann--Whitney).
The reach rate of \ours ($95$--$98\%$) is below that of \noapf
($99$--$100\%$) at every radius, as under hidden observation.
The training curves ($n=20$ seeds) are
consistent with this: the mean collision rate of \ours reaches zero at the
$9\times10^4$-step checkpoint and ends at zero, whereas \noapf still collides in
$13.1\%$ of episodes at $1.5\times10^5$ steps, and $15$ of the $20$ \noapf seeds
are above zero at the end of training.
Providing the obstacle features therefore gives \noapf no benefit that is
visible at this sample size. A plausible explanation is that, with a collision penalty
of only $0.2$ per step and a short path rewarded by the step cost, the obstacle
features alone do not make a detour profitable, but we did not test this explanation.
The APF advantage therefore cannot be attributed to the hidden observation.

\subsubsection{Off-policy solvers}\label{sec:exp-solver}

\begin{figure}[!t]
\centering
\includegraphics[width=\linewidth]{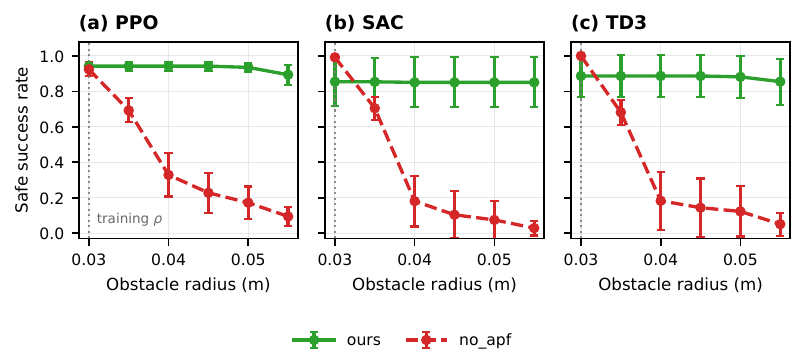}
\caption{Safe success rate vs.\ obstacle radius for (a)~PPO, (b)~SAC, and (c)~TD3, each
panel comparing \ours against \noapf ($n=30$ seeds for PPO, $n=20$ for SAC
and TD3; error bars denote $95\%$ confidence intervals).}\label{fig:solver}
\end{figure}

To test whether the effect is specific to PPO, we retrain \ours and
\noapf with SAC~\cite{haarnoja2018} and TD3~\cite{fujimoto2018} in the same environment, with the same reward, training
budget, and evaluation ($n=20$ seeds each; observations flattened to a $12$-dimensional
Box for these off-policy solvers).
SAC uses two hidden layers of $64$ units, and TD3 two of $128$ units with Gaussian
exploration noise ($\sigma=0.3$); both use a batch size of $64$ and one gradient update
every two environment steps.
Because \ours and \noapf share these settings within each solver, the
comparison between them is not confounded by the solver-specific choices.
Seeds~1--10 were trained first. Two TD3 runs of \ours among them (seeds~3 and~10)
had not converged under this setting and were retrained with one gradient update per
environment step; no \noapf run was retrained. Seeds~11--20 were added later; none of
them was retrained, and all of their runs are reported as trained.
Table~\ref{tab:solver} reports the safe success rate, and Figure~\ref{fig:solver} compares it with that of PPO.
\begin{table}[htbp]
\centering
\caption{Safe success rate (\%) for SAC and TD3 ($n=20$ seeds each).}\label{tab:solver}
\begin{tabular}{llcccccc}
\toprule
Solver & Group & \multicolumn{6}{c}{Obstacle radius $\rho$ (m)}\\
\cmidrule(lr){3-8}
& & 0.030 & 0.035 & 0.040 & 0.045 & 0.050 & 0.055\\
\midrule
SAC & \ours & 85.5 & 85.4 & 85.1 & 85.1 & 85.1 & 85.1\\
SAC & \noapf & 99.2 & 70.5 & 18.1 & 10.4 & 7.5 & 2.9\\
TD3 & \ours & 88.7 & 88.7 & 88.7 & 88.7 & 88.2 & 85.5\\
TD3 & \noapf & 100.0 & 68.2 & 18.3 & 14.4 & 12.3 & 5.1\\
\bottomrule
\end{tabular}
\end{table}

At the training radius, \noapf reaches a higher safe success rate than \ours
for both solvers ($99.2\%$ vs.\ $85.5\%$ for SAC, Welch $p=0.048$; $100\%$ vs.\ $88.7\%$
for TD3, $p=0.064$). This gap is not caused by collisions but by seeds of \ours
that did not learn to reach the goal: two seeds per solver (SAC seeds~17 and~18, TD3 seeds~15 and~20)
reach the goal in fewer than $50\%$ of the episodes. Among the other $18$ seeds, the
safe success rate of \ours is $94.6\%$ for SAC at every radius and $96.4\%$ for TD3
up to $\rho=0.045\,\mathrm{m}$ ($92.9\%$ at $\rho=0.055\,\mathrm{m}$). This subset
is a post hoc selection and is not used for any claim; it only indicates where the gap
originates.
At every OOD radius, \noapf collapses while \ours stays nearly flat. The OOD
advantage is $+14.9$ to $+82.2$ percentage points for SAC and $+20.5$ to $+80.4$ percentage
points for TD3. For $\rho\ge0.04\,\mathrm{m}$ the Welch tests give $p<2.7\times10^{-8}$ for SAC
and $p<1.8\times10^{-8}$ for TD3; at $\rho=0.035\,\mathrm{m}$ they give $p=0.049$ for SAC and
$p=0.0044$ for TD3. The Mann--Whitney tests give $p<4.2\times10^{-4}$ (SAC) and
$p<7.3\times10^{-4}$ (TD3) at every OOD radius.
This addresses the
concern that an off-policy algorithm might learn a safety margin without the APF term:
the clearance of \noapf at the training radius remains small for both solvers
($7.8\,\mathrm{mm}$ for SAC, $8.3\,\mathrm{mm}$ for TD3, against $54.0\,\mathrm{mm}$
and $64.3\,\mathrm{mm}$ for \ours). This is expected, because the no-APF reward
does not depend on the clearance as long as the EE stays outside the pillar, while the
step cost favors the shortest collision-free path.

\section{Discussion}\label{sec:discussion}

The experiments indicate that, among the reward terms tested, the horizontal
artificial potential field is the one responsible for size robustness under hidden
observation, rather than the goal-progress reward or the vertical ceiling penalty.
The remainder of this section interprets the result, states its cost, and delimits
the claims.

\subsection{Interpretation of the Mechanism}\label{sec:disc-interp}

The robustness rests on a distinction between two objectives that a safety
signal can encode. A signal that penalizes only penetration, as the collision
indicator of Equation~\eqref{eq:collision} does, is flat with respect to
clearance everywhere outside the pillar surface. Combined with a step cost, it
favors the shortest path, which skims the surface and leaves almost
no margin for a radius that increases after training. The horizontal potential of
Equation~\eqref{eq:apf} instead moves the penalty boundary outward to the safety
margin $d_{\mathrm{safe}}$, so the optimum it selects at the training radius is a
clearance of roughly $47.9\,\mathrm{mm}$, versus $9.6\,\mathrm{mm}$ without the term
(Table~\ref{tab:margin}).

The policy does not observe the pillars, so the executed trajectory depends on the
goal alone and the minimum clearance decreases with radius at a slope of exactly $-1$
(Section~\ref{sec:exp-margin}). The policy holds a center distance, $d_c + \rho$, of
$77.9\,\mathrm{mm}$ for \ours and $75.8\,\mathrm{mm}$ for \noprog,
versus $38.4$--$39.7\,\mathrm{mm}$ for the three groups without the term
(Table~\ref{tab:margin}). In both cases the distance is an absolute distance from the
pillar center, because the radius is not observed. The term therefore does not make
the policy adapt to the radius; it enlarges the buffer between the training surface
and the trajectory. The buffer of $47.9\,\mathrm{mm}$ exceeds the largest radius
increase of the sweep ($25\,\mathrm{mm}$), which is why the mean clearance stays
positive over the tested range, and the $5.3\%$ collision rate of \ours at
$\rho=0.055\,\mathrm{m}$ shows that the buffer does not suffice for every seed and goal.

This observation bears on the relation between learned and analytical safety, but
the present experiments do not settle it. A learned safety signal that is flat inside
the safe region encodes only the boundary encountered during training, whereas an
analytical term can specify a preferred distance from that boundary. What the
experiments show is that this specification yields a larger buffer. A larger collision
penalty is the natural alternative, and over the range tested it did not reproduce the
buffer: penalties of $5$ and $10$ leave a safe success rate of $8.8\%$ and $18.0\%$ at
$\rho=0.055\,\mathrm{m}$, against $89.4\%$ for \ours
(Section~\ref{sec:exp-collpen}). Only two penalty values were tested, so this does not
show that no penalty can produce the buffer, but it indicates that a buffer does not
follow simply from making contact more expensive.
The comparison with the classical APF planner (Section~\ref{sec:exp-classic}) adds
a second perspective: the same field used as a planner, with access to the obstacle
state, attains zero collision and a $90\%$ safe success rate at every radius and adapts
its clearance to the radius, at the cost of a path that is $2.1$ to $2.6$ times longer.
The field used as a reward transfers part of this safety to a policy that does not
observe the obstacles.

\subsection{The Cost of the Margin and the Over-the-Top Route}\label{sec:disc-cost}

The margin is not cost-free. The APF groups follow trajectories of roughly $0.57$
and $0.59\,\mathrm{m}$, versus $0.48\,\mathrm{m}$ for the groups without the
term, because the step cost forces the policy to trade path length against
safety and the wider detour is the equilibrium of that trade-off. A second cost is
visible in the safe success rate of $94.3\%$ for \ours: with a collision rate of
zero at the training radius, the remaining $5.7\%$ of episodes end without reaching
the goal and without collision. These episodes concentrate at the far edge of the goal
region ($84\%$ of goals with $x<-0.355\,\mathrm{m}$ are reached, against $98\%$
elsewhere), whereas \noapf reaches both parts at the same rate. We did not
establish the cause of this shortfall. The margin should therefore be interpreted as a
conservatism rather than as a cost-free gain, and its magnitude is set by the relative
weights of the potential and the step cost, which we did not vary.

A natural objection is that the policy could avoid this trade-off altogether by
passing above the pillars, which do not reach the workspace ceiling. No trained
policy does so: the over-top rate is zero for every group at every radius,
including the groups whose reward contains no ceiling penalty
(Section~\ref{sec:exp-setup}). To test whether this reflects return
maximization rather than an artifact, we searched $75$ scripted over-top
trajectories per goal and compared the best collision-free one with the learned
policies under the same reward. The low-altitude route attains a higher return
for all three groups and is surpassed by an over-top trajectory on only one to
three of the twelve goals, so the low route is consistent with return
maximization under our reward. We do not claim this to be a global optimum: the
search covers a restricted, three-parameter trajectory family, and the height
penalty of Equation~\eqref{eq:height} is a soft guard that raises the cost of
the over-top route without forbidding it. The preference for the low route is
therefore reported as an empirical finding, not as a design guarantee.

\subsection{Sample Efficiency and the Budget-Limited Claim}\label{sec:disc-eff}

The auxiliary term also changes the rate at which collisions disappear during
training. The APF groups reach and hold a collision rate of zero at a median of
$1.5\times10^4$ steps, whereas $21$ of the $30$ \noapf seeds do so within
the budget, at a median of $1.2\times10^5$ steps, and the other nine remain above zero
at the $1.5\times10^5$-step budget (Section~\ref{sec:exp-efficiency}). The speedup
concerns collisions only: the safe success rate during training does not differ
significantly between \ours and \noapf at most checkpoints.
We do not claim that the APF term makes safe behavior
unreachable in the limit. At the training radius the collision penalty and the
potential point toward the same optimum, so a longer run may eliminate collisions even
without the term, and the difference in safe success rate there is not significant
($p=0.44$; Section~\ref{sec:exp-significance}). The claim is instead twofold. First, the
potential makes collision-free behavior reachable within a practical budget, which
matters whenever the budget, rather than the asymptotic performance, is the binding
constraint. Second, and more important, the optimum it selects has a larger margin, and
only the larger margin remains collision-free at out-of-distribution radii. A longer
budget might let \noapf eliminate collisions at the training radius, but the
trajectory it converges to would still sit near the pillar surface
(Table~\ref{tab:margin}).

\subsection{Limitations}\label{sec:disc-limits}

We identify eight limitations of the present study.

\begin{enumerate}
\item \textbf{Simulation domain.} The results are obtained in PyBullet through
\emph{panda-gym} on a single reaching task, with kinematic end-effector control, no
contact dynamics, and non-terminal collisions in which the end-effector may pass
through a pillar. Transfer to a real manipulator must therefore account for perception
noise, full dynamics, and collision that is both physical and terminal; we make no
claim about deployment safety beyond the abstraction of the simulator.

\item \textbf{Fixed layout.} The nominal pillar centers are identical in every
training and evaluation episode; hidden observation withholds only the
$\pm15\,\mathrm{mm}$ drift and the radius, not the layout itself. The policy can
therefore store the layout in its weights, and the results describe a fixed workcell
with an uncertain obstacle size. They do not cover a layout that changes between
episodes, for which a policy that does not observe the obstacles has no basis for
avoidance, nor a larger or unstructured position uncertainty.

\item \textbf{Obstacle model.} The potential acts in the horizontal plane on the
end-effector position against two cylindrical pillars, complemented by a soft
ceiling penalty, and the collision check ignores the arm links. Generalizing
the approach to arbitrary obstacle shapes, full-arm collision, or
three-dimensional surfaces requires a signed distance that respects the actual
geometry and a retrained policy, which we leave to future work.

\item \textbf{Geometric range.} The radius sweep is bounded above by
$0.055\,\mathrm{m}$ because the corridor between the nominal pillar surfaces closes at
$\rho\approx0.062\,\mathrm{m}$, and enlarging the pillar separation would make
the straight line collision-free and remove the discrimination the task is
designed to produce (Section~\ref{sec:exp-setup}). The out-of-distribution range
we probe is therefore structural rather than arbitrary, and robustness beyond it
remains untested.

\item \textbf{Fixed margin and limited hyperparameter study.} The buffer is a fixed
distance determined by the training radius, $d_{\mathrm{safe}}$, and $\lambda_s$; it does
not adapt to the radius at test time, so the collision rate of \ours rises from
$0.9\%$ at $\rho=0.05\,\mathrm{m}$ to $5.3\%$ at $\rho=0.055\,\mathrm{m}$ as the buffer
is consumed. The default coefficients were set once, before the experiments. We varied
$d_{\mathrm{safe}}$ and $\lambda_s$ one at a time around the default ($10$ seeds per
setting, \ref{app:sensitivity}): the clearance and the path length
grew with both, the safe success rate at $\rho=0.055\,\mathrm{m}$ fell for a smaller
margin, and $\lambda_s=3$ prevented $3$ of $10$ seeds from learning to reach the goal.
The default is not claimed to be optimal, and the study does not cover interactions
between the two coefficients.

\item \textbf{Collision semantics.} Collisions are non-terminal and carry a
per-step penalty of $0.2$, a compromise adopted because terminal collisions
collapse the no-APF policy to a $100\%$ collision rate during training
(Section~\ref{sec:reward-collision}). This compromise weakens the collision
signal: \noapf prefers to penetrate rather than to detour as the radius
grows, so the steep decline in its safe success rate partly reflects that
penetration is inexpensive under this reward. The reported collision rates
therefore quantify safety under this reward rather than physical safety. A larger
penalty did not compensate for this. With penalties of $5$ and $10$ ($25\times$ and
$50\times$ the default; $10$ seeds each, Section~\ref{sec:exp-collpen}), the mean
clearance was $9.9$ and $17.3\,\mathrm{mm}$, and only $1$ and $2$ of the $10$ seeds
kept a safe success rate above $50\%$ at $\rho=0.055\,\mathrm{m}$, against all $30$
seeds of \ours. We did not test larger penalties or other solvers, so this result
shows that the potential is more reliable than a larger penalty over the range tested,
not that no penalty can reproduce it.

\item \textbf{Safe-RL baseline.} We compare against reward ablations, an
analytic planner, and three solvers, but not against methods that enforce safety
explicitly, such as constrained policy optimization~\cite{achiam2017}, Lagrangian
methods, or control-barrier-function filters. Constraint-based methods that need only
a cost signal could be trained under hidden observation, and whether they produce a
clearance beyond the training radius, alone or with a clearance-dependent cost, is
untested here. Filters that act on the obstacle state or model need it at run time,
so a full-observation safe reinforcement learning baseline is a natural next step for
testing whether the advantage also holds when such a filter can observe the obstacle.
We also did not train with the radius randomized, which would remove the
out-of-distribution premise of the sweep.

\item \textbf{Sample sizes and solver stability.} The main experiments use $30$ seeds per
group, the full-observation, SAC, and TD3 experiments use $20$, and the
collision-penalty control uses $10$ per penalty. Two of the $20$ SAC and two of the $20$
TD3 runs of \ours did not learn to reach the goal. No seed was excluded, and the only
reruns were the two TD3 runs of \ours described in Section~\ref{sec:exp-solver}, so these failures lower the safe success rate of \ours at the training radius
below that of \noapf for both solvers. The classical planner is deterministic and is evaluated on $50$ goals, so no
confidence interval applies to it. The agreement in direction across observation models
and solvers is therefore more reliable than the size of the individual differences.
\end{enumerate}

\section{Conclusion}\label{sec:conclusion}

We studied safe reaching under hidden observation in a fixed workcell, where the
policy receives no information about the instantaneous obstacle position, motion,
or radius, and asked how reward design can make the resulting policy robust to
out-of-distribution obstacle radii.
Augmenting the reward with an analytic artificial potential field, active in the
horizontal plane within a safety margin around each pillar, moved the learned
optimum from a near-zero clearance to a positive one. The minimum clearance at the
training radius is $47.9\,\mathrm{mm}$ with the term versus $9.6\,\mathrm{mm}$ without
it, and this buffer is what carries over to larger radii: the APF
policies maintain a safe success rate above $88\%$ throughout the sweep, whereas the
policies without the term fall below $10\%$ at the maximum tested radius, a
difference that is statistically significant at every out-of-distribution radius. The auxiliary term also eliminates collisions during
training at least $8\times$ earlier in median training steps. The effect persists under
full observation and across three RL solvers (PPO, SAC, and TD3), and the resulting
policy, which does not observe the obstacles, attains a safe success rate comparable to
that of a classical APF planner that reads the pillar positions.

The margin is a fixed buffer rather than an adaptation to the radius, because the
policy never observes the radius. It exceeds the radius increases tested here, and
the collision rate of \ours becomes non-zero only at the two largest radii.
The buffer is not free: the trajectory is longer ($0.57\,\mathrm{m}$ versus
$0.48\,\mathrm{m}$ without the term), and $5.7\%$ of episodes at the training
radius end without reaching the goal. Raising the collision penalty by $25\times$ or $50\times$ left only $1$ and $2$
of $10$ seeds with a safe success rate above $50\%$ at the largest radius, against all
$30$ seeds with the potential, so over the range tested the potential is a more
reliable source of margin than a larger penalty.

Future work proceeds along three directions: transfer to a real manipulator, where
perception noise, full dynamics, and physical, terminal collision must be accounted
for; extension of the potential to general obstacle geometry and full-arm collision,
replacing the planar clearance with a signed distance to the arm links; and a
comparison with dedicated safe reinforcement learning methods under full observation,
to test whether the hidden-observation advantage also holds when a safety filter has
access to the obstacle.


\section*{CRediT authorship contribution statement}
\textbf{Ziyi Hao:} Conceptualization, Methodology, Software, Validation,
Formal analysis, Investigation, Writing -- original draft, Writing -- review
\& editing.

\section*{Declaration of competing interest}
The author declares that he has no known competing financial interests or
personal relationships that could have appeared to influence the work reported
in this paper.

\section*{Declaration of generative AI and AI-assisted technologies in the manuscript preparation process}
During the preparation of this work the author used DeepSeek to assist with the
development of the simulation and analysis code, and Claude (Anthropic) to polish
the language of the manuscript. After using these tools, the author reviewed and
edited the content as needed and takes full responsibility for the content of the
published article.

\section*{Funding}
This research did not receive any specific grant from funding agencies in the
public, commercial, or not-for-profit sectors.

\section*{Data availability}
The code, the trained models, the per-seed evaluation data, and the script that
recomputes the reported numbers (\texttt{verify\_paper\_numbers.py}) are available at
\url{https://github.com/Hao-Ziyi/apf-reward-margin} under the MIT license.
A permanent archive of the released version is deposited in Zenodo~\cite{hao2026code}
(\url{https://doi.org/10.5281/zenodo.23885305}).

\clearpage 
\appendix
\counterwithin{table}{section}
\counterwithin{algocf}{section}
\counterwithin{figure}{section}
\renewcommand{\thetable}{\Alph{section}.\arabic{table}}
\renewcommand{\thealgocf}{\Alph{section}.\arabic{algocf}}
\renewcommand{\thefigure}{\Alph{section}.\arabic{figure}}

\section{Reward pseudocode}\label{app:algorithm}

Algorithm~\ref{alg:apf-reward} evaluates the reward of Equation~\eqref{eq:reward}
for one environment step. It restates the terms of Table~\ref{tab:reward} and
Equations~\eqref{eq:smoothness}--\eqref{eq:collision} in the order in which they are computed.

\begin{algorithm}[H]
\caption{APF-augmented reward for one environment step}
\label{alg:apf-reward}
\SetKwInOut{Input}{Input}
\SetKwInOut{Output}{Output}
\Input{end-effector position $\mathbf{p}$, previous position $\mathbf{p}_{\mathrm{prev}}$,
action $\mathbf{a}$, previous action norm $\lVert \mathbf{a}_{\mathrm{prev}} \rVert$,
goal $\mathbf{g}$, pillars $\{(\mathbf{c}_i, \rho, h)\}_{i=1}^{N}$}
\Output{reward $r$, collision flag $c$}
$d \leftarrow \lVert \mathbf{p} - \mathbf{g} \rVert$\;
$d_{\mathrm{prev}} \leftarrow \lVert \mathbf{p}_{\mathrm{prev}} - \mathbf{g} \rVert$\;
$r \leftarrow -d$\;
\If{$d < \delta$}{$r \leftarrow r + \lambda_b$\;}
$r \leftarrow r + \lambda_p \max(0,\, d_{\mathrm{prev}} - d)$\;
$r \leftarrow r - \lambda_\sigma \left( \lVert \mathbf{a} \rVert - \lVert \mathbf{a}_{\mathrm{prev}} \rVert \right)^{2}$\;
$\Phi \leftarrow 0$\; $c \leftarrow \mathrm{false}$\;
\ForEach{pillar $(\mathbf{c}_i, \rho, h)$}{
    $d_{c,i} \leftarrow \lVert \mathbf{p}_{xy} - \mathbf{c}_{i,xy} \rVert - \rho$\;
    \If{$0 \le z \le h$ \textbf{and} $d_{c,i} < d_{\mathrm{safe}}$}{
        $\Phi \leftarrow \Phi + \left( (d_{\mathrm{safe}} - \max(d_{c,i},\, 0)) / d_{\mathrm{safe}} \right)^{2}$\;
    }
    \If{$d_{c,i} < 0$ \textbf{and} $0 \le z \le h$}{$c \leftarrow \mathrm{true}$\;}
}
$r \leftarrow r - \lambda_s \Phi$\;
$\bar{z} \leftarrow \max(0,\, z - h)$\;
$r \leftarrow r - \lambda_h \left( \bar{z}/0.04 + (\bar{z}/0.04)^{2} \right)$\;
\If{$c$}{$r \leftarrow r - \lambda_c$\;}
$r \leftarrow r - \lambda_t$\;
\Return{$r$, $c$}\;
\end{algorithm}

\section{Implementation and training details}\label{app:implementation}

Table~\ref{tab:impl} collects the settings that are shared by all experiments or that
define the solvers and the classical planner. Settings that are not listed use the
defaults of Stable-Baselines3 2.9.0~\cite{raffin2021}.

\begin{table}[H]
\centering
\small
\caption{Environment, solver, and planner settings.}\label{tab:impl}
\begin{tabular}{@{}lp{0.68\linewidth}@{}}
\toprule
Item & Setting\\
\midrule
Environment & \texttt{PandaReach-v3} (dense reward), $100$ steps per episode\\
Action & $[-1,1]^3$, scaled to a maximum step of $0.05\,\mathrm{m}$\\
Success threshold & $0.03\,\mathrm{m}$; discount factor $\gamma=0.99$\\
Pillars & $N=2$, height $0.21\,\mathrm{m}$, nominal centers $(-0.07, 0.045)$ and
$(-0.19, 0.015)\,\mathrm{m}$; random walk along $x$ with a step drawn uniformly from
$[-5,5]\,\mathrm{mm}$, reflected at $\pm15\,\mathrm{mm}$ from the nominal center; reset to the nominal
centers at every episode\\
Training radius & $\rho=0.03\,\mathrm{m}$\\
Goal sampling & $x\in[-0.38,-0.30]$, $y\in[0.03,0.15]$, $z\in[0.08,0.13]\,\mathrm{m}$,
rejected if within $\rho+0.03\,\mathrm{m}$ of a nominal pillar center\\
Training budget & $1.5\times10^5$ environment steps per run\\
\midrule
PPO & \texttt{MultiInputPolicy}, entropy coefficient $0.01$, other settings default\\
SAC & \texttt{MlpPolicy}, two hidden layers of $64$ units, batch size $64$, one gradient
update every two environment steps, automatic entropy coefficient\\
TD3 & \texttt{MlpPolicy}, two hidden layers of $128$ units, batch size $64$, one gradient
update every two environment steps (one update per step for the retrained runs of seeds~3 and~10, Section~\ref{sec:exp-solver}), Gaussian exploration noise $\sigma=0.3$, policy delay $2$\\
SAC and TD3 input & $12$-dimensional vector: end-effector state ($6$), goal ($3$), achieved goal ($3$)\\
\midrule
Classical APF & attraction gain $K_{\mathrm{att}}=20$ per meter of goal offset in $x$, $y$, and $z$;
repulsion $K_{\mathrm{rep}}(1/d_c-1/d_0)/d_c^{2}$ with $K_{\mathrm{rep}}=5$ and
$d_0=0.08\,\mathrm{m}$, active for $0<d_c<d_0$ and $0\le z\le h$; resultant command
saturated at unit norm\\
\bottomrule
\end{tabular}
\end{table}

\section{Over-the-top route check}\label{app:overtop}

The pillars do not reach the workspace ceiling, so an end-effector trajectory above
$z=0.21\,\mathrm{m}$ passes them without collision. No trained policy exploits this
route: the over-top rate, the fraction of episodes in which the end-effector is above a
pillar footprint, is $0\%$ for every policy at every radius, including \noapf and
\vanilla, whose rewards contain no ceiling penalty.

To test whether this is consistent with return maximization, we searched $75$ scripted
over-top trajectories per goal on $12$ goals. Each trajectory is parameterized by a
cruise height, a rise point, and a descent point. We compared the best collision-free
trajectory with the learned policies (PPO, seeds~1--5), which pass below the pillar
tops, under the same reward. For \noapf at $\rho=0.03\,\mathrm{m}$, the best over-top
trajectory attains a mean return of $2.24$ against $3.27$ for the learned policy and
needs $11.3$ instead of $9.8$ steps on average. The ordering is the same for
\vanilla ($-3.03$ vs.\ $-2.27$) and \ours ($0.89$ vs.\ $1.65$), and the over-top
trajectory exceeds the learned return on only one to three of the $12$ goals. The
low-altitude route is therefore consistent with return maximization under this reward,
although the search covers a restricted trajectory family and is not a proof of global
optimality.

In addition, in $500$ evaluation episodes per group (PPO, seeds~1--10, $50$ goals) the
end-effector height never exceeds its starting value of $0.197\,\mathrm{m}$, which is
$13\,\mathrm{mm}$ below the pillar top, so the learned trajectories stay clear of the
region where the ceiling penalty applies.

\section{Sensitivity to the potential parameters}\label{app:sensitivity}

The default coefficients $d_{\mathrm{safe}}=0.08\,\mathrm{m}$ and $\lambda_s=1.5$ were
fixed before the experiments of Section~\ref{sec:experiments}. To see how the learned
margin depends on them, we changed one coefficient at a time and trained \ours with
PPO under hidden observation at the training radius $\rho=0.03\,\mathrm{m}$ ($10$ seeds
per setting, seeds~1--10, otherwise the protocol of Table~\ref{tab:impl}). The policies
were evaluated on the radius sweep of Section~\ref{sec:exp-main}.
Table~\ref{tab:sens} reports the results. The default row is computed from the same
seeds~1--10 of the main experiment, so its values differ slightly from the $30$-seed
values reported there; \noapf on the same seeds has a clearance of $10.1\,\mathrm{mm}$ and
a safe success rate of $11.6\%$ at $\rho=0.055\,\mathrm{m}$.

\begin{table}[!htbp]
\centering
\small
\setlength{\tabcolsep}{3.5pt}
\caption{Sensitivity of \ours to $d_{\mathrm{safe}}$ and $\lambda_s$ ($n=10$ seeds per row).
Clearance and path length are measured at the training radius $\rho=0.03\,\mathrm{m}$;
the safe success rate and the collision rate are percentages. The last column counts the
seeds whose safe success rate at $\rho=0.055\,\mathrm{m}$ is at least $50\%$.}\label{tab:sens}
\begin{tabular}{@{}ccccccccc@{}}
\toprule
$d_{\mathrm{safe}}$ & $\lambda_s$ & Clearance & Path & \multicolumn{2}{c}{Safe success (\%)} & Collision (\%) & Seeds\\
\cmidrule(lr){5-6}
(mm) & & (mm) & (m) & $\rho=0.03$ & $\rho=0.055$ & $\rho=0.055$ & $\geq50\%$\\
\midrule
$40$ & $1.5$ & $25.3$ & $0.522$ & $91.6$ & $53.2$ & $42.2$ & $9/10$\\
$60$ & $1.5$ & $36.8$ & $0.528$ & $93.2$ & $81.8$ & $12.6$ & $10/10$\\
$80$ (default) & $1.5$ & $46.5$ & $0.581$ & $95.2$ & $89.8$ & $5.8$ & $10/10$\\
$100$ & $1.5$ & $56.5$ & $0.600$ & $92.0$ & $84.2$ & $9.2$ & $9/10$\\
$80$ & $0.75$ & $25.0$ & $0.515$ & $91.0$ & $51.8$ & $42.8$ & $7/10$\\
$80$ & $3.0$ & $67.8$ & $0.764$ & $55.6$ & $54.4$ & $1.4$ & $7/10$\\
\bottomrule
\end{tabular}
\end{table}

Three observations follow. First, the clearance grows with both coefficients, from
$25\,\mathrm{mm}$ to $57\,\mathrm{mm}$ over $d_{\mathrm{safe}}=40$--$100\,\mathrm{mm}$ and from
$25\,\mathrm{mm}$ to $68\,\mathrm{mm}$ over $\lambda_s=0.75$--$3$, and the path grows with it.
Second, a smaller margin ($d_{\mathrm{safe}}=40\,\mathrm{mm}$ or $\lambda_s=0.75$) leaves a
clearance of about $25\,\mathrm{mm}$, and the collision rate at $\rho=0.055\,\mathrm{m}$ rises to
about $42\%$, against $5.8\%$ for the default. The values of $d_{\mathrm{safe}}$ of
$60$ and $100\,\mathrm{mm}$ did not differ significantly from the default in safe success
rate at $\rho=0.055\,\mathrm{m}$ (Welch $p=0.27$ and $0.56$). Third, $\lambda_s=3$ almost
removes collisions, but it prevented $3$ of the $10$ seeds from learning to reach the goal
(reach rate at most $8\%$ at the training radius), so its safe success rate stays near $55\%$ at
every radius. Every setting exceeded \noapf in safe success rate at $\rho=0.055\,\mathrm{m}$
(Welch $p<0.01$ for all five), so the positive margin does not depend on the default
values. The study uses $10$ seeds per setting without correction for multiple
comparisons, and it does not show that the default is optimal.

\clearpage 
%
%

\end{document}